\documentclass{article}

\PassOptionsToPackage{table}{xcolor}
\usepackage{arxiv}
\usepackage[utf8]{inputenc}
\usepackage[T1]{fontenc}
\usepackage{hyperref}
\usepackage{url}
\usepackage{booktabs}
\usepackage{amsfonts}
\usepackage{nicefrac}
\usepackage{microtype}
\usepackage{xcolor}
\usepackage{amsmath}
\usepackage{graphicx}
\usepackage{multirow}
\usepackage{caption}
\usepackage{array}
\usepackage{subcaption}
\usepackage{overpic}
\usepackage{float}
\usepackage{collcell}
\usepackage{pgf}
\usepackage{wrapfig}
\usepackage{ifthen}

\title{AEGIR: Modeling Area Emitters for Indoor Inverse Rendering using Gaussian Splatting}

\author{
\begingroup
\renewcommand{\arraystretch}{1.16}
\setlength{\tabcolsep}{1.85em}
\begin{tabular}{@{}cc@{}}
{\large Mohamed Shawky Sabae} & {\large Philipp Langsteiner} \\[-0.05em]
{\small\texttt{mohamed.sabae@uni-tuebingen.de}} &
{\small\texttt{philipp.langsteiner@uni-tuebingen.de}} \\[1.05em]
{\large Jan-Niklas Dihlmann} & {\large Hendrik Lensch} \\[-0.05em]
{\small\texttt{jan-niklas.dihlmann@uni-tuebingen.de}} &
{\small\texttt{hendrik.lensch@uni-tuebingen.de}}
\end{tabular}
\endgroup
\\[3.0em]
{\normalsize University of T\"ubingen}
\\
}
\date{June 2026}

\renewcommand{\headeright}{A Preprint}
\renewcommand{\undertitle}{A Preprint}
\renewcommand{\shorttitle}{AEGIR}

\hypersetup{
  colorlinks=true,
  linkcolor=black,
  citecolor=black,
  urlcolor=black,
  pdftitle={AEGIR: Modeling Area Emitters for Indoor Inverse Rendering using Gaussian Splatting},
  pdfauthor={Mohamed Shawky Sabae, Philipp Langsteiner, Jan-Niklas Dihlmann, Hendrik Lensch}
}
\AtBeginDocument{%
  \newgeometry{%
    textheight=8.72in,
    textwidth=6.10in,
    top=1in,
    headheight=14pt,
    headsep=25pt,
    footskip=30pt
  }%
}

\makeatletter
\renewcommand{\section}{%
  \@startsection{section}{1}{\z@}%
                {-2.2ex \@plus -0.5ex \@minus -0.2ex}%
                { 1.8ex \@plus  0.3ex \@minus  0.2ex}%
                {\large\bf\raggedright}%
}
\renewcommand{\subsection}{%
  \@startsection{subsection}{2}{\z@}%
                {-2.0ex \@plus -0.5ex \@minus -0.2ex}%
                { 1.05ex \@plus  0.2ex}%
                {\normalsize\bf\raggedright}%
}
\renewcommand{\subsubsection}{%
  \@startsection{subsubsection}{3}{\z@}%
                {-1.7ex \@plus -0.5ex \@minus -0.2ex}%
                { 0.75ex \@plus  0.2ex}%
                {\normalsize\bf\raggedright}%
}
\renewcommand{\@toptitlebar}{%
  \hrule height 2\p@
  \vskip 0.18in
  \vskip -\parskip%
}
\renewcommand{\@bottomtitlebar}{%
  \vskip 0.16in
  \vskip -\parskip
  \hrule height 2\p@
  \vskip 0.08in%
}
\renewcommand{\@maketitle}{%
  \vbox{%
    \hsize\textwidth
    \linewidth\hsize
    \vskip 0.06in
    \@toptitlebar
    \centering
    {\LARGE\sc \@title\par}
    \@bottomtitlebar
    \textsc{\undertitle}\\
    \vskip 0.08in
    \def\And{%
      \end{tabular}\hfil\linebreak[0]\hfil%
      \begin{tabular}[t]{c}\bf\rule{\z@}{20\p@}\ignorespaces%
    }
    \def\AND{%
      \end{tabular}\hfil\linebreak[4]\hfil%
      \begin{tabular}[t]{c}\bf\rule{\z@}{20\p@}\ignorespaces%
    }
    \begin{tabular}[t]{c}\bf\rule{\z@}{22\p@}\@author\end{tabular}%
    \vskip 0.12in \@minus 0.05in \center{\@date} \vskip 0.13in
  }%
}
\makeatother

\begin{document}

\maketitle

\begin{abstract}
Inverse rendering requires separating illumination from surface materials, which is highly ambiguous due to their tight coupling in observed images. While Gaussian Splatting is efficient for novel view synthesis, existing relightable methods approximate scene lighting using discrete point lights, global environment maps, or implicit representations. By ignoring the physical spatial extent of real-world emitters, these approaches produce incorrect light attenuation and unrealistic shadows. We present \textbf{AEGIR} (\textbf{A}rea \textbf{E}mitters for \textbf{G}aussian \textbf{I}nverse \textbf{R}endering), a framework that explicitly models local area emitters within a relightable Gaussian Splatting representation. Joint optimization of emitters, materials, and geometry is challenging due to flexible emitter parameterization, which increases both the number of parameters and the ambiguity between illumination and materials. We address this by introducing a differentiable deferred rendering pipeline that integrates multiple importance sampling with targeted regularization. As a result, AEGIR accurately simulates local light transport and achieves more consistent decomposition. Experiments show that explicit area emitters improve illumination reconstruction and enhance downstream tasks, including novel view synthesis, controlled relighting, and virtual object insertion, particularly in scenes with complex local lighting. Project page: \href{https://darkgeekms.github.io/projects/aegir}{\textcolor{blue}{https://darkgeekms.github.io/projects/aegir}}.
\end{abstract}

\section{Introduction}\label{sec:intro}
\begin{figure}[t]
    \centering
    \includegraphics[width=0.94\linewidth]{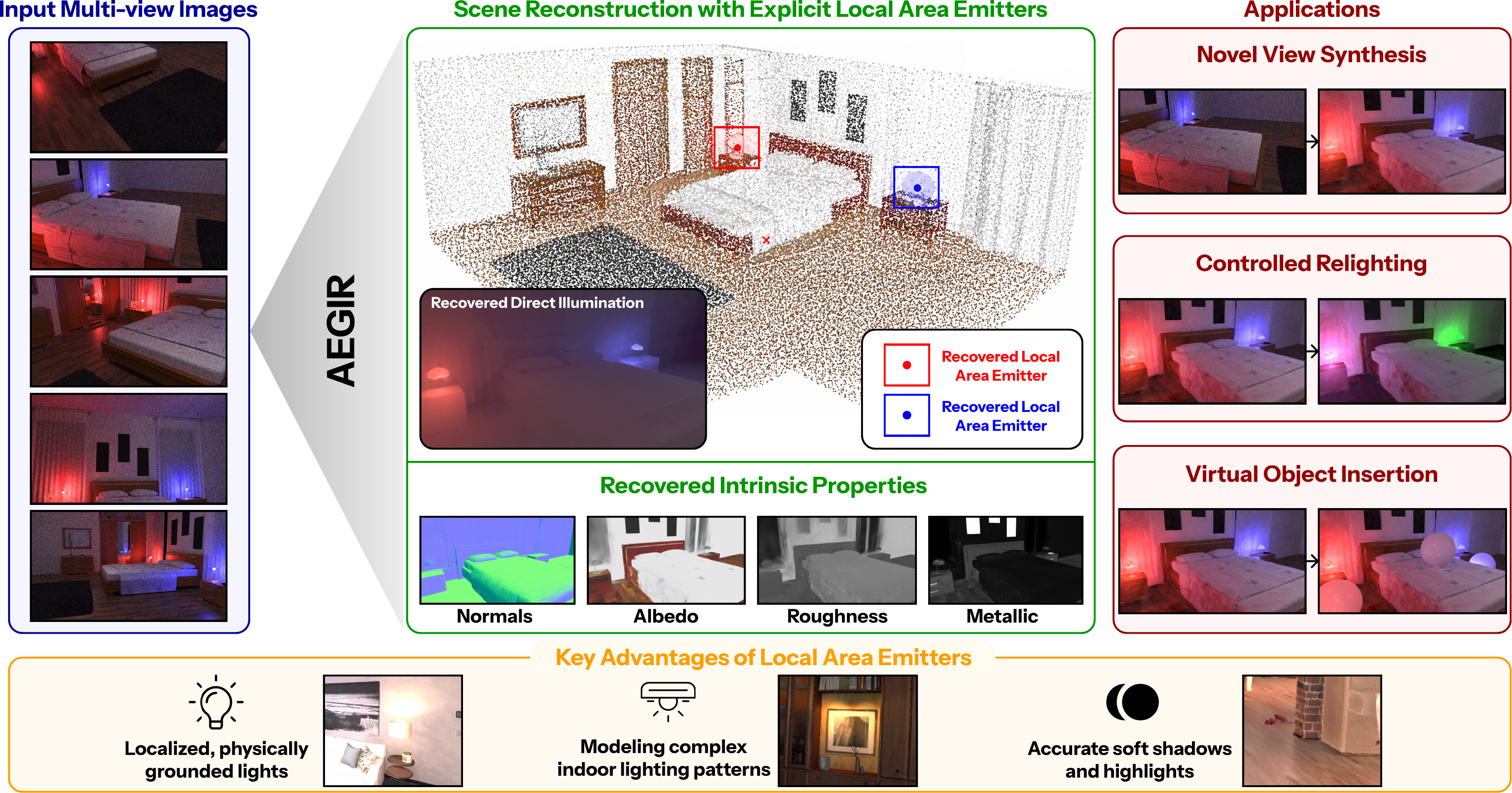}
    \caption{\textbf{AEGIR Overview.} From multi-view images, AEGIR jointly models local area emitters, 3D geometry, and materials. These explicit emitters provide localized, physically grounded lighting, capturing accurate shadows and complex lighting patterns for realistic novel view synthesis, scene relighting, and virtual object insertion.}
    \label{fig:teaser}
\end{figure}

Reconstructing 3D environments from 2D images has advanced rapidly using Neural Radiance Fields (NeRF) \cite{mildenhall2020nerf} and 3D Gaussian Splatting (3DGS) \cite{kerbl3Dgaussians}. Although highly effective for novel view synthesis, extending these representations to inverse rendering remains a challenge. Inverse rendering seeks to decompose a scene into geometry, materials, and illumination, but this problem is inherently ill-posed because appearance depends on the combined effects of surface material and lighting. This decomposition is essential for downstream tasks, such as scene editing and relighting.

A core limitation in existing inverse rendering approaches lies in the representation of illumination. Existing approaches typically rely on global environment maps \cite{R3DG2023, gu2024IRGS}, discrete point lights \cite{kocsis2024iid, du2025gsidilluminationdecompositiongaussian}, or implicit neural representations \cite{yao2022neilf, zhang2023neilf++}. These models struggle in bounded indoor environments because real-world emitters, such as lamps and ceiling panels, have finite spatial extents and close proximity to the scene. Environment maps assume infinite distance, and point lights lack the physical dimensions to cast soft shadows. Although implicit light fields capture spatial position, their lack of explicit geometry makes it difficult to enforce strict physical constraints like attenuation over distance. By ignoring the physical dimensions of local emitters, these approximations produce incorrect light attenuation, unrealistic illumination patterns, and inaccurate shadow boundaries. Consequently, the optimization often absorbs residual lighting effects directly into the albedo.

We address these limitations with AEGIR (Figure \ref{fig:teaser}), a framework that explicitly models area light emitters within a GS representation. AEGIR parameterizes lighting as localized 3D primitives with spatial positions, shapes, and angular emission profiles. Leveraging anisotropic scaling and Super-Gaussian angular falloff distributions, our representation mathematically adapts to reconstruct a wide range of emitters. This flexibility allows a single primitive to seamlessly transition from compact point light sources to elongated fluorescent tubes, providing a more expressive and physically meaningful approximation of real illumination.

However, area emitters are difficult to optimize due to their high-dimensional parameterization, which amplifies the ambiguity between material and illumination. By enforcing radiometric and geometric constraints and introducing a physically grounded, fully differentiable parameterization, our model enables stable optimization while accurately capturing localized light transport. AEGIR jointly optimizes these area emitters alongside scene geometry and physically based (PBR) material parameters, specifically diffuse albedo, roughness, and metallic, directly from multi-view images. Experiments demonstrate that our emitter representation improves illumination estimation and downstream tasks, including novel view synthesis and editing in scenes with complex local lighting.

In summary, our core contributions are:
\begin{itemize}
    \setlength{\itemsep}{1pt}
    \setlength{\parsep}{0pt}
    \setlength{\parskip}{0pt}
    \setlength{\topsep}{2pt}
    \item A novel explicit area emitter formulation that parameterizes localized 3D primitives with shapes and emission profiles. By combining anisotropic scaling with a Super-Gaussian angular falloff, this representation enables flexible modeling of diverse real-world emitters.
    \item A joint optimization GS framework that simultaneously recovers these explicit light sources, scene geometry, and PBR materials directly from multi-view images, achieving proper disentanglement between material and lighting using radiometric and geometric constraints.
    \item A physically grounded rendering model that evaluates illumination directly from area emitters, enabling physically consistent downstream applications such as novel view synthesis, controlled scene relighting, and virtual object insertion.
\end{itemize}

\section{Related Work}\label{sec:related}
\subsection{Inverse Rendering and Intrinsic Decomposition}

\textbf{Data-Driven Inverse Rendering.} 
Data-driven approaches learn priors from large-scale datasets to predict intrinsic scene properties from images. Early neural methods \cite{yu19inverserendernet, li2019inverserenderingcomplexindoor, zhu2022irisformerdensevisiontransformers} have been increasingly replaced by diffusion-based models such as IntrinsicDiffusion \cite{Luo2024IntrinsicDiffusion}, RGB$\leftrightarrow$X \cite{zeng2024rgb}, and DiffusionRenderer \cite{DiffusionRenderer}, which estimate intrinsic attributes including albedo, roughness, and surface normals. To improve multi-view consistency, recent methods such as IDArb \cite{li2025idarb} and MatSpray \cite{langsteiner2025matsprayfusing2dmaterial} leverage diffusion priors, while MVInverse \cite{wu2025mvinversefeedforwardmultiviewinverse} adopts a feed-forward architecture with explicit consistency constraints. Although these approaches produce high-quality intrinsic predictions, they often lack explicit physically accurate modeling of light transport.

\textbf{Optimization-Based Inverse Rendering.} 
Optimization-based methods enforce multi-view consistency by minimizing photometric error using differentiable rendering. NeRF-based approaches such as NeRD \cite{boss2021nerd}, NeRFactor \cite{zhang2021nerfactor}, Neural-PIL \cite{boss2021neuralpil}, NeILF \cite{yao2022neilf}, and NeILF++ \cite{zhang2023neilf++} jointly estimate geometry, materials, and illumination, but are computationally expensive. More recent 3DGS formulations, including Relightable 3D Gaussian \cite{R3DG2023}, GS-IR \cite{liang2024gsir3dgaussiansplatting}, SVG-IR \cite{sun2025svgirspatiallyvaryinggaussiansplatting}, IRGS \cite{gu2024IRGS}, GeoSplatting \cite{ye2025geosplatting}, and GS-ID \cite{du2025gsidilluminationdecompositiongaussian}, integrate physically based rendering (PBR) for efficient relighting. Methods such as FIPT \cite{fipt2023} and IRIS \cite{lin2025iris} improve the decomposition of indoor scenes by using structured lighting representations and stronger geometric and material priors.

Existing methods rely on simplified illumination models, such as environment maps \cite{Munkberg_2022_CVPR, hasselgren2022nvdiffrecmc, R3DG2023, yao2025refGS, gu2024IRGS}, discrete point lights (often modeled as Spherical Gaussians) \cite{kocsis2024iid, du2025gsidilluminationdecompositiongaussian, niu2026sgsintrinsicsemanticinvariantgaussiansplatting}, and implicit representations \cite{yao2022neilf, zhang2023neilf++, 10.1111:cgf.15014, zhu2022learning}, which ignore the physical spatial extent of real-world emitters. These approaches introduce ambiguity in inverse rendering, often leading to residual lighting effects being absorbed into the albedo. Motivated by this, AEGIR explicitly models localized geometric area lights within Gaussian Splatting, enabling more accurate and physically grounded illumination decomposition.

\subsection{Illumination Modeling in 3D Environments}

Accurate illumination modeling is essential for inverse rendering and realistic image synthesis, as it governs how light interacts with scene materials. A common representation is to model lighting using global environment maps, often parameterized by Spherical Gaussians \cite{physg2021}. While effective for distant illumination, these models assume spatially uniform lighting and struggle to capture localized variation and distance attenuation in indoor scenes. Nevertheless, environment maps remain widely used, estimated using feedforward neural networks \cite{Garon_2019_CVPR, li2023spatiotemporallyconsistenthdrindoor} or, more recently, latent diffusion models \cite{Phongthawee2023DiffusionLight, 10.1145/3721238.3730749, liang2025luxdit}. To introduce spatial locality, methods such as Deep Parametric Indoor Lighting \cite{Gardner_2019_ICCV} regress discrete 3D light sources from a single image. Although they provide spatially localized illumination, they model emitters as isotropic area lights, which cannot represent the spatial structure of real light sources and lack temporal consistency. Hybrid approaches, such as GS-ID \cite{du2025gsidilluminationdecompositiongaussian} and SGS-Intrinsic \cite{niu2026sgsintrinsicsemanticinvariantgaussiansplatting}, model illumination by combining global environment maps with localized point lights represented as Spherical Gaussians. Alternatively, recent generative methods such as LumiNet \cite{xing2025luminet} and LuxRemix \cite{liang2026luxremixlightingdecompositionremixing} leverage diffusion models to decompose complex indoor lighting into editable components for relighting. Despite these differences, both families of methods lack explicit modeling of spatially extended emitters, which limits their ability to model physically accurate lighting effects.

In contrast, AEGIR explicitly parameterizes the spatial extents, shapes, and angular emission profiles of light emitters under proper optimization constraints. This physically grounded formulation enables more accurate illumination decomposition and improves disentanglement between materials and lighting compared to prior global, point-based, hybrid, and purely generative approaches.

\section{Methodology}\label{sec:method}
\begin{figure}[t]
    \centering
    \includegraphics[width=\linewidth]{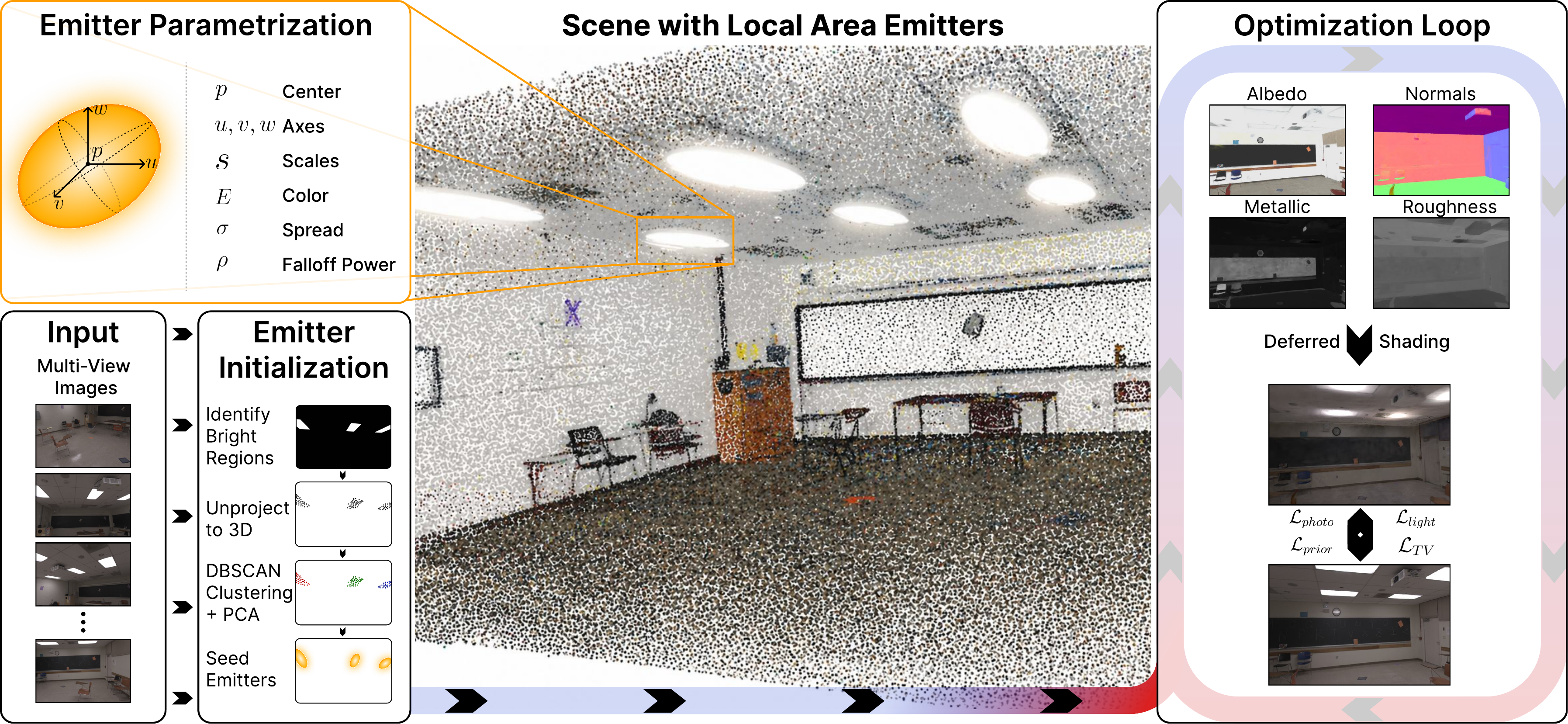}
    \caption{\textbf{AEGIR Optimization Framework.} Parameterized area emitters are initialized from multi-view inputs and jointly optimized with 2D Gaussian geometry and PBR materials within a deferred rendering framework, enabling physically grounded scene decomposition.}
    \label{fig:pipeline}
\end{figure}

To model complex spatially varying illumination, we propose a unified area emitter representation that explicitly parameterizes light shape and emission profile to simulate realistic attenuation and soft occlusions. As illustrated in Figure \ref{fig:pipeline}, by coupling these emitters with relightable 2D Gaussians and evaluating illumination through a deferred rendering framework with multiple importance sampling, we jointly optimize illumination and intrinsic materials under carefully designed constraints. This results in physically grounded illumination and consistent material disentanglement.

\subsection{Preliminaries: Relightable Gaussian Splatting}

Standard Gaussian Splatting \cite{kerbl3Dgaussians} reconstructs scenes by optimizing volumetric primitives whose colors are baked into view-dependent Spherical Harmonics (SH). To enable dynamic relighting, Relightable Gaussian Splatting \cite{R3DG2023} decouples illumination from appearance, replacing SH coefficients with explicit microfacet Bidirectional Reflectance Distribution Function (BRDF) parameters (typically diffuse albedo and roughness) that physically respond to novel lighting conditions. However, computing accurate physically based shading requires precise surface normals. Therefore, we rely on 2D Gaussian Splatting (2DGS) \cite{Huang2DGS2024}, whose planar Gaussian representation produces more reliable surface geometry and normals than volumetric splatting. This makes 2DGS well-suited for stable physically based shading.

\subsection{Explicit Local Area Emitters}

\paragraph{Anisotropic Super-Gaussian Emitter Formulation.} We propose representing explicit area emitters as a dynamic collection of anisotropic 3D ellipsoids with a Super-Gaussian angular falloff distribution. Each emitter $e$ is parameterized by its spatial center $p_e \in \mathbb{R}^3$, local orthonormal axes $\{\mathbf{u}_e,\mathbf{v}_e,\mathbf{w}_e\}$, physical scales $\mathbf{S}_e=(S_{e,u},S_{e,v},S_{e,w}) \in \mathbb{R}_+^3$, RGB emission color $\mathbf{E}_e \in \mathbb{R}_+^3$, axis-aligned angular spread coefficients $\boldsymbol{\sigma}_e=(\sigma_{e,u},\sigma_{e,v},\sigma_{e,w}) \in \mathbb{R}_+^3$, and a Super-Gaussian angular falloff power $\rho_e \in \mathbb{R}_+$. For compact notation, $\mathbf{k}_e$ denotes the corresponding emitter axis for $k\in\{u,v,w\}$, i.e., $\mathbf{u}_e$, $\mathbf{v}_e$, or $\mathbf{w}_e$.

For a given surface point $x$, let $\omega_{ie}$ denote the incident light direction pointing from the surface toward emitter $e$. We formulate the radiance $\mathbf{L}_{ie}(x, \omega_{ie})$ from an explicit emitter as an anisotropic Super-Gaussian angular emission profile:
\begin{equation}
    \mathbf{L}_{ie}(x, \omega_{ie}) =
    \mathbf{E}_e \,
    \exp \left(
    - \left(
    \sum_{k\in\{u,v,w\}}
    \frac{(\omega_{ie} \cdot \mathbf{k}_e)^2}{\sigma_{e,k}^2}
    \right)^{\rho_e}
    \right)
\label{eq:light_radiance}
\end{equation}
We model light transport using the Monte Carlo estimator in Equation \ref{eq:app_monte_carlo} (Appendix \ref{append:render}), combining BRDF samples with samples drawn from the explicit emitters. Light samples are generated by mapping unit sphere samples to the surface of each anisotropic ellipsoid, as shown in Equation \ref{eq:app_light_surface_sample}. The corresponding BRDF Probability Distribution Function (PDF) and light sampling surrogate are given in Equations \ref{eq:app_pdf_brdf} and \ref{eq:app_pdf_light}, and are combined using the Multiple Importance Sampling (MIS) balance heuristic in Equation \ref{eq:app_mis_balance}. Because the light sampling surrogate depends on the distance between the surface point and the sampled emitter point, distance attenuation is captured in the direct shading estimate.

\begin{figure}[t]
    \centering
    \begin{subfigure}[b]{0.48\linewidth}
        \includegraphics[width=\linewidth]{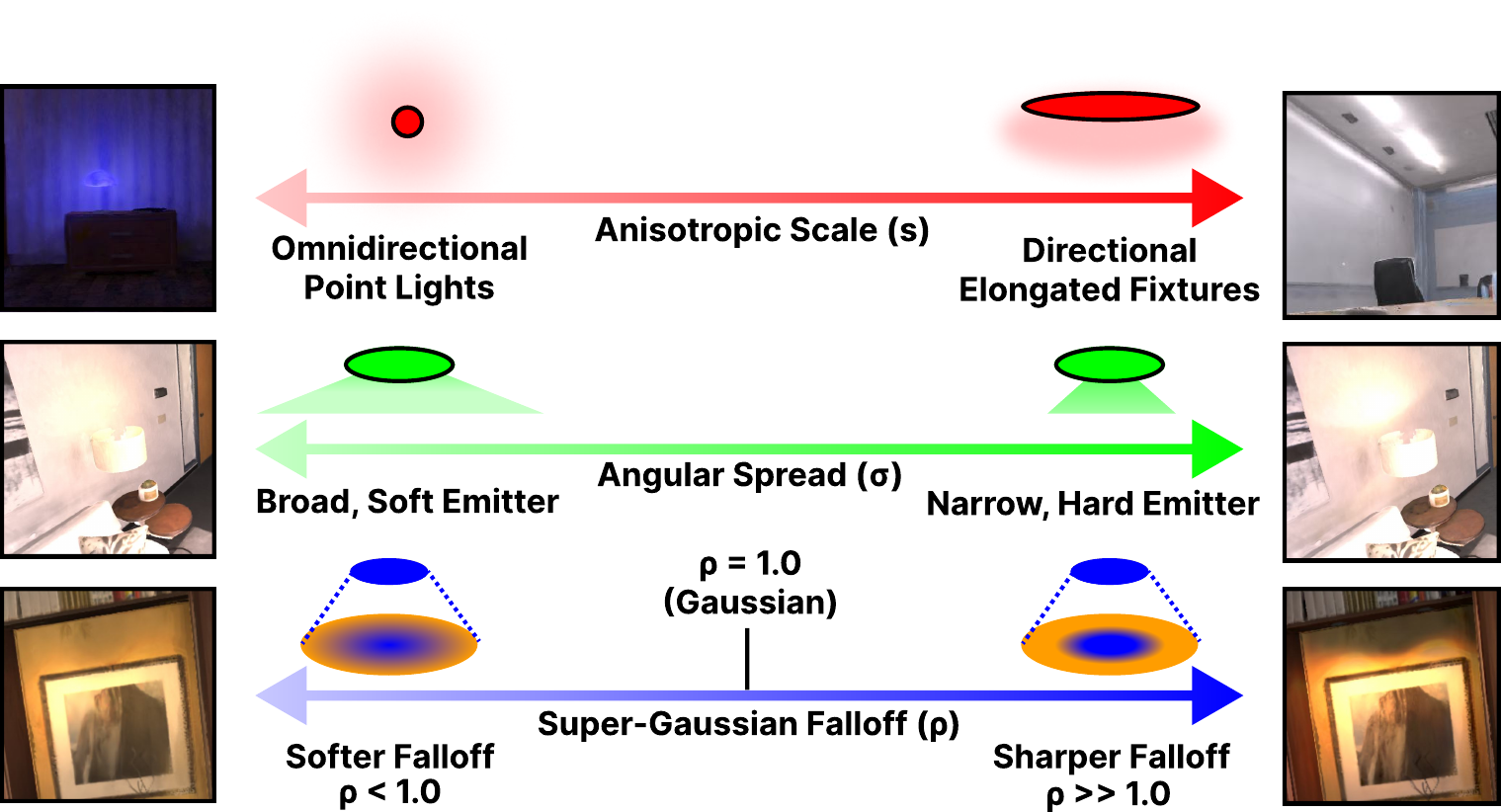}
        \caption{Effect of area emitter parameters.}
        \label{fig:emitter_param_effect}
    \end{subfigure}
    \hfill 
    \begin{subfigure}[b]{0.48\linewidth}
        \includegraphics[width=\linewidth]{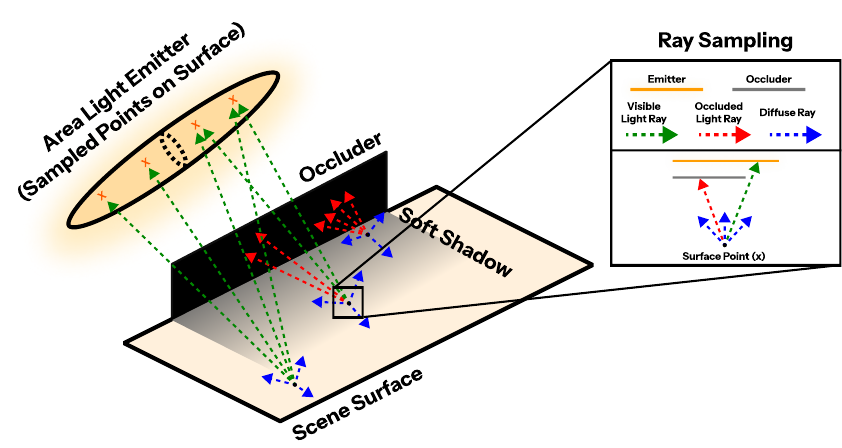}
        \caption{Area emitter sampling and visibility tracing.}
        \label{fig:emitter_sampling}
    \end{subfigure}
    
    \caption{\textbf{Area Emitter Formulation and Evaluation.} (a) Emitters use scales ($\mathbf{S}_e$), angular spread ($\boldsymbol{\sigma}_e$), and Super-Gaussian angular falloff ($\rho_e$). (b) Illumination is evaluated using emitter surface sampling and visibility tracing. This physically grounded ray-tracing approach resolves complex occlusions and generates accurate soft shadows.}
    \label{fig:combined_light}
\end{figure}

To understand how Equation \ref{eq:light_radiance} achieves both complex anisotropy and variable sharpness, we can decompose the angular spread mechanism into two primary components (illustrated in Figure \ref{fig:emitter_param_effect}):

\begin{enumerate}
    \item \textbf{Axis-Aligned Anisotropic Spread:} Rather than uniform angular spread, we project the incident light direction $\omega_{ie}$ onto the emitter's local coordinate frame $\{\mathbf{u}_e,\mathbf{v}_e,\mathbf{w}_e\}$. By weighting these projections with independent learnable angular spread coefficients $\sigma_{e,u}$, $\sigma_{e,v}$, and $\sigma_{e,w}$, we decouple light spread across different axes. This anisotropic scaling allows a single primitive to represent a wide range of physical light sources, spanning from near-isotropic point lights to the directional elongated footprints of fixtures such as fluorescent tubes.

    \item \textbf{Super-Gaussian Angular Falloff:} The inclusion of a learnable power parameter $\rho_e$ elevates the standard Gaussian distribution to a Super-Gaussian distribution. While a standard formulation ($\rho_e=1$) results in smooth bell-curved falloff with soft penumbras, raising the anisotropic angular distance to the power of $\rho_e$ allows the model to manipulate the light source falloff. For rays emitted near the central axis, a high $\rho_e$ maintains a near-uniform intensity. However, as the emission angle increases, the high exponent causes the radiance to drop abruptly, creating a sharper falloff. This enables the model to differentiate between bare, omnidirectional bulbs (low $\rho_e$) and sharply housed, directional spotlights (high $\rho_e$).
\end{enumerate}

 \paragraph{Emitter Initialization and Adaptive Control.} To constrain the ill-posed joint optimization, we initialize area emitters by unprojecting high-intensity RGB pixels into a 3D point cloud. These points are clustered using DBSCAN \cite{10.5555/3001460.3001507} to identify physical light fixtures, with each cluster used to set the initial center, scale, and axes of an emitter using Principal Component Analysis (PCA). During optimization, we update all parameters and dynamically manage the emitter count. The adaptive controller uses the perceptual emitter energy $\Phi_e$ defined in Equation \ref{eq:app_emitter_energy} (Appendix \ref{append:implement}). Lights with negligible energy or those drifting out of bounds are pruned, while those exceeding the energy or scale thresholds in Appendix \ref{append:implement} are split along their largest-scale axis to refine localized illumination. Let $k_e^\star=\arg\max_{k\in\{u,v,w\}}S_{e,k}$ and let $\mathbf{k}_e^\star\in\{\mathbf{u}_e,\mathbf{v}_e,\mathbf{w}_e\}$ denote the corresponding axis. For a split, the centers of the children $p_{e,a}$ and $p_{e,b}$ are translated from the parent center $p_e$ as:
\begin{equation}
p_{e,a}, p_{e,b} = p_e \pm \frac{1}{2} S_{e,k_e^\star} \mathbf{k}_e^\star
\label{eq:emitter_split}
\end{equation}


\subsection{Deferred Rendering of Relightable Gaussians and Area Emitters}

AEGIR utilizes deferred rendering to decouple geometry and material rasterization from lighting evaluation. We rasterize 2D Gaussians into a G-buffer containing albedo, roughness, metallic, and surface normals, and then evaluate the rendering equation with a microfacet BRDF using the Monte Carlo estimator in Equation \ref{eq:app_monte_carlo}. To reduce variance, we employ MIS, illustrated in Figure \ref{fig:emitter_sampling}, combining cosine-weighted hemisphere sampling for diffuse transport (Equation \ref{eq:app_pdf_brdf}) with direct light sampling for high frequency shadows and specular highlights. For direct lighting, we sample each explicit emitter using the ellipsoid surface mapping in Equation \ref{eq:app_light_surface_sample}, then combine these samples with BRDF samples using the light sampling surrogate and MIS weights in Equations \ref{eq:app_pdf_light} and \ref{eq:app_mis_balance}. Rays are then traced through the 2D Gaussian scene, as shown in Figure \ref{fig:emitter_sampling}, to compute a soft differentiable visibility term $V_i(x, \omega_i)$:
\begin{equation}
V_i(x, \omega_i) = 1 - \alpha_{trace} \cdot \operatorname{sigmoid} \left( d_{light} - \epsilon - d_{trace} \right)
\label{eq:visibility_term}
\end{equation}
where $\alpha_{trace}$ is the accumulated opacity along the ray, $d_{light}$ is the distance from the surface to the sampled emitter point, $d_{trace}$ is the traced occlusion distance, and $\epsilon$ is a bias term for stability. Subsequently, the total incident light is obtained by weighting direct emission $L_i^{direct}(x, \omega_i)$ and a single-bounce indirect approximation $L_i^{bounce}(x, \omega_i)$:
\begin{equation}
L_i(x, \omega_i) = V_i(x, \omega_i) \cdot L_i^{direct}(x, \omega_i)  + (1 - V_i(x, \omega_i)) \cdot L_i^{bounce}(x, \omega_i)
\label{eq:combined_light}
\end{equation}
 We model indirect illumination with a single secondary bounce. During training, $L_i^{bounce}$ is approximated with the SH/color features of the secondary Gaussian hit for faster and more stable optimization. However, during inference, we sample a diffuse direction at the secondary hit, trace it through the scene, and evaluate emission only if the ray reaches one of the recovered explicit emitters, otherwise the bounce contribution is zero. This gives a cheap one-bounce estimate rather than a full multi-bounce path tracer. Finally, the total incident light $L_i(x, \omega_i)$ gathered from all samples is integrated against the microfacet BRDF to evaluate the final pixel color. Detailed mathematical formulations for the microfacet BRDF model and the deferred rendering pipeline are provided in Appendix \ref{append:render}.


\subsection{Joint Optimization and Training Curriculum}

Decomposing observed radiance into geometry, materials, and lighting is a highly ambiguous problem. This difficulty increases when modeling area light emitters, as they introduce more unknown variables. Without strict constraints, the optimization tends to bake lighting effects, such as shadows or specular highlights, into the material representation, degrading both disentanglement and illumination estimation. To resolve these ambiguities, AEGIR employs a progressive curriculum. We first establish a geometric baseline using DN-Splatter \cite{turkulainen2024dnsplatter}. This phase computes a dense 2D Gaussian geometry and initial Spherical Harmonics (SH), which are retained only as a training-time proxy for secondary-bounce illumination in the subsequent joint optimization step. Once the baseline is established, we jointly optimize the Gaussian geometry, microfacet materials, and area emitters. This phase is driven by a combined objective function consisting of photometric, material, and emitter-specific constraints:
\begin{equation}
\mathcal{L}_{total} = \mathcal{L}_{photo} + \mathcal{L}_{prior} + \mathcal{L}_{TV} + \mathcal{L}_{light}
\label{eq:total_loss}
\end{equation}

\textbf{Contrast-Aware Photometric Loss ($\mathcal{L}_{photo}$):} We introduce a focus map $W_f$ that upweights pixels with high local contrast. This forces the optimization to prioritize the reconstruction of high-frequency lighting cues that standard losses often blur.

\textbf{Diffusion Priors ($\mathcal{L}_{prior}$):} To resolve global ambiguities such as distinguishing a white wall in shadow from a dark wall, we supervise materials with pseudo-ground truth from a pretrained DiffusionRenderer \cite{DiffusionRenderer}. This prevents the model from erroneously adjusting the albedo to compensate for complex lighting.

\textbf{Edge-Aware Total Variation ($\mathcal{L}_{TV}$):} To mitigate noise in the diffusion priors, we apply a total variation (TV) loss on the material maps guided by ground-truth image gradients. This encourages piecewise constant materials on flat surfaces while preserving sharp structural boundaries, forcing the emitters rather than the material to explain illumination patterns.

\textbf{Emitter Regularization ($\mathcal{L}_{light}$):} To keep explicit lights physically plausible, we penalize excessive spatial scaling, encourage color neutrality to prevent illumination from bleeding into the albedo, and enforce bounds on the spatial position to keep emitters within the observable scene.

Detailed mathematical formulations for all loss components are provided in Appendix \ref{append:loss}. Finally, we freeze the scene geometry and materials, optimizing only the area emitters. This forces the lighting representation to resolve any remaining discrepancies, allowing us to refine soft shadows and structured highlights without baking illumination into the albedo.

\section{Results and Discussion}\label{sec:results}
\subsection{Experimental Setup and Datasets}

We evaluate our method on synthetic environments with complex illumination (Hypersim \cite{roberts:2021}, FIPT \cite{fipt2023}) and real-world indoor scans (Replica \cite{replica19arxiv}, ScanNet++ \cite{yeshwanth2023scannet++}). We compare against GS-ID \cite{du2025gsidilluminationdecompositiongaussian}, IRGS \cite{gu2024IRGS}, IRIS \cite{lin2025iris}, and NeILF++ \cite{zhang2023neilf++}. Since the official GS-ID implementation is not publicly available, we reproduce their hybrid illumination formulation and deshadowing model for evaluation. After geometry initialization, we optimize AEGIR's deferred shading stage for 20,000 iterations using the Adam optimizer on a single NVIDIA RTX 4090 GPU, which takes approximately 40 minutes. Unless otherwise stated, quantitative image metrics are computed on the raw rendered outputs before any denoising. The OptiX denoiser \cite{10.1145/1778765.1778803} is used only when saving final visualization images to reduce Monte Carlo noise. Detailed dataset splits, hyperparameters, sampling counts, and regularization weights are provided in Appendix \ref{append:implement}.


\subsection{Core Contribution: Explicit Local Area Emitters}

\begin{figure*}[ht]
    \centering
    
    \newlength{\imgwidth}
    \setlength{\imgwidth}{0.23\textwidth} 
    \newlength{\insetwidth}
    \setlength{\insetwidth}{0.09\textwidth} 
    
    \newcommand{\headerbox}[1]{%
        \makebox[\imgwidth][c]{\scriptsize #1}%
    }
    
    \newcommand{\withinset}[2]{%
        \IfFileExists{#1}{%
            \begin{overpic}[width=\imgwidth]{#1}
                \put(-3,-3){%
                    \setlength{\fboxsep}{0pt}%
                    \setlength{\fboxrule}{0.3pt}%
                    \IfFileExists{#2}{%
                        \fcolorbox{red}{white}{\includegraphics[width=\insetwidth]{#2}}%
                    }{%
                        \fcolorbox{red}{white}{\parbox[c][0.5\insetwidth][c]{\insetwidth}{\centering\fontsize{4}{5}\selectfont \textcolor{black}{N/A}}}%
                    }%
                }
            \end{overpic}%
        }{%
            \setlength{\fboxsep}{0pt}%
            \setlength{\fboxrule}{0.4pt}%
            \framebox{\parbox[c][0.75\imgwidth][c]{\imgwidth}{\centering \small \textbf{N/A}}}%
        }%
    }
    
    \setlength{\tabcolsep}{1.5pt} 
    \renewcommand{\arraystretch}{1.0} 
    
    \begin{tabular}{@{} c @{\hspace{3mm}} c @{}}
        
        & \begin{tabular}{@{} c c c c @{}}
            \headerbox{Reference} & 
            \headerbox{\textbf{AEGIR (Ours)}} & 
            \headerbox{GS-ID \cite{du2025gsidilluminationdecompositiongaussian}} & 
            \headerbox{IRGS \cite{gu2024IRGS}}
          \end{tabular} \\ [2mm]
        
        \rotatebox[origin=c]{90}{\small \textbf{Indoor}} &
        \begin{tabular}{@{} c c c c @{}}
            \includegraphics[width=\imgwidth]{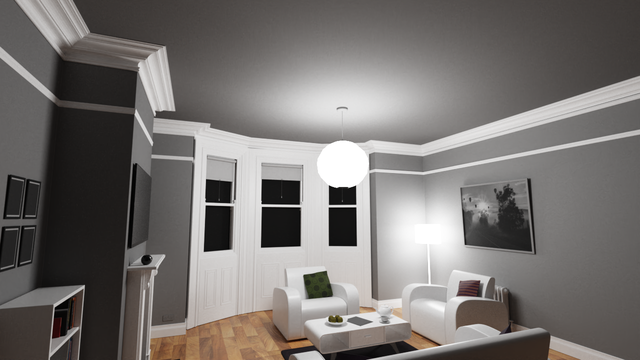} &
            \withinset{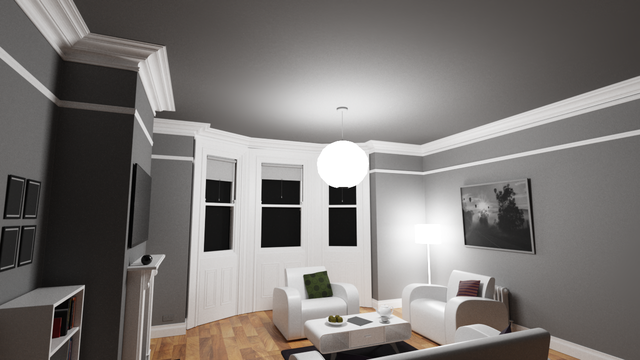}{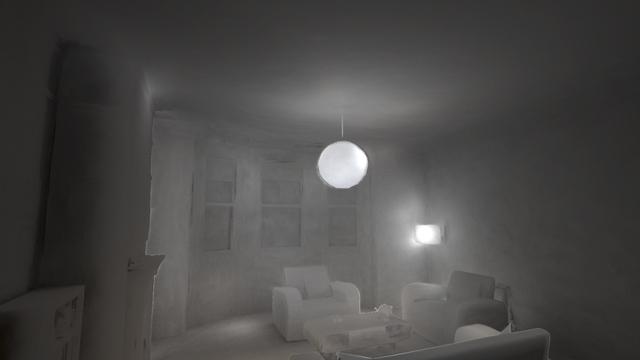} &
            \withinset{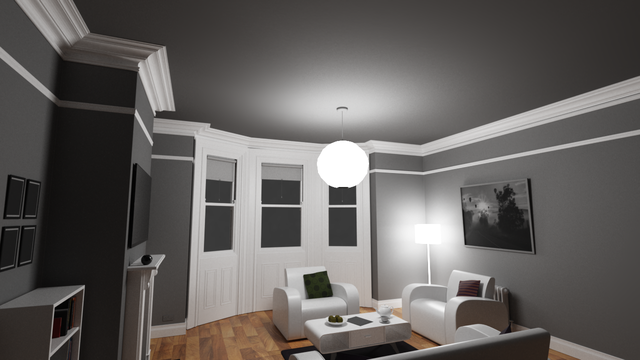}{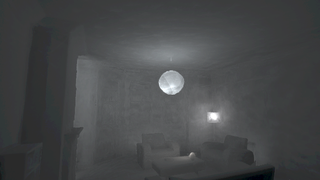} & 
            \withinset{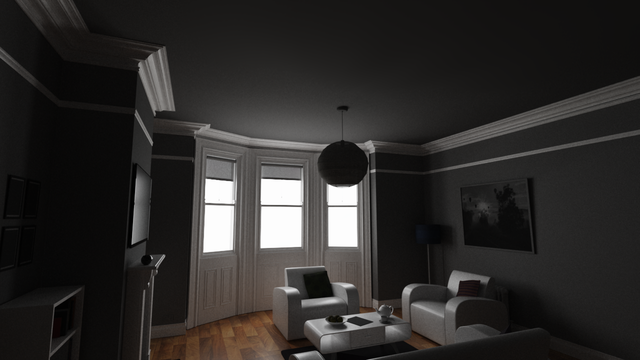}{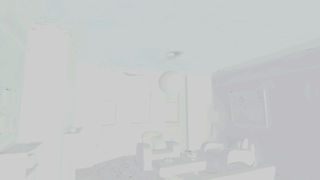} \\ [1.5mm]
        \end{tabular} \\ [4mm] 
        
        \rotatebox[origin=c]{90}{\small \textbf{Environment}} &
        \begin{tabular}{@{} c c c c @{}}
            \includegraphics[width=\imgwidth]{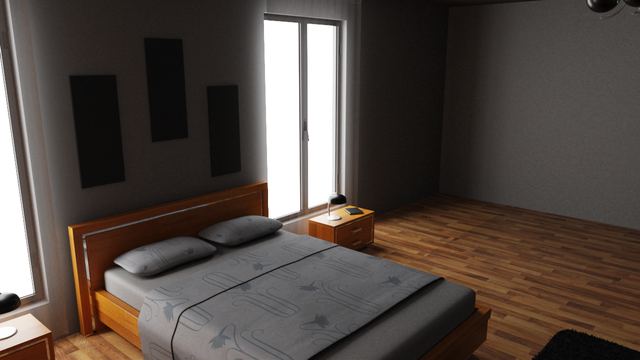} &
            \withinset{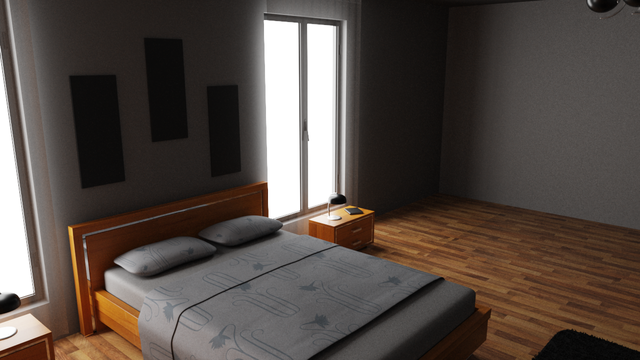}{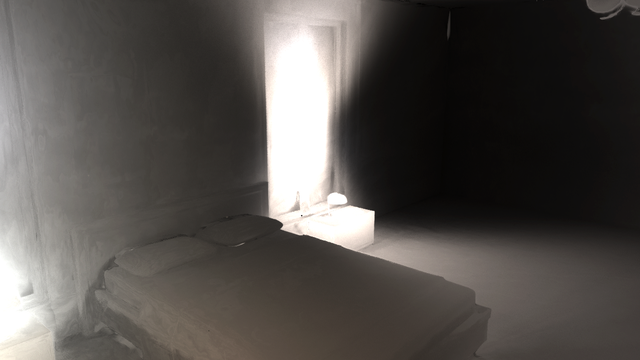} &
            \withinset{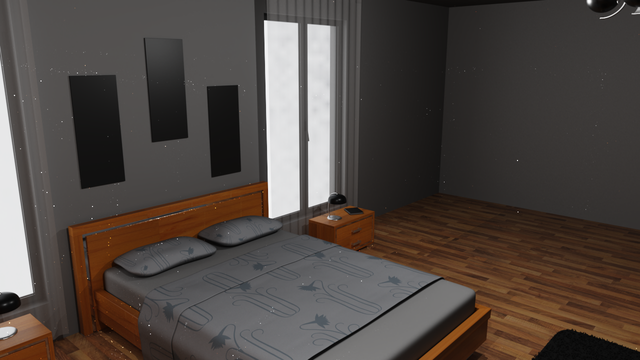}{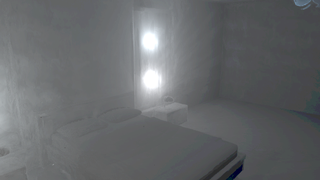} & 
            \withinset{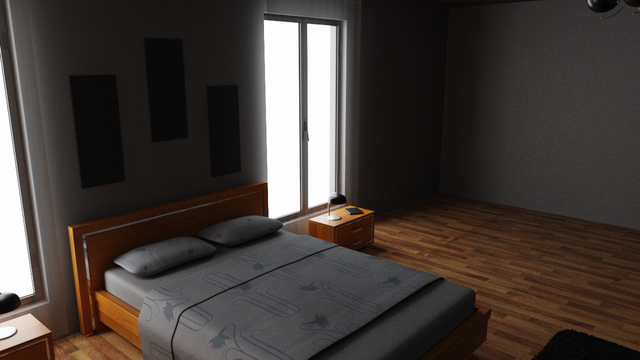}{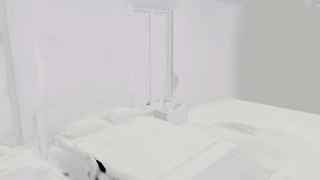} \\
        \end{tabular} \\
        
    \end{tabular}
    
    \vspace{0.5mm}
    \caption{\textbf{Qualitative comparison of pure lighting estimation.} All images are rendered with Mitsuba \cite{Mitsuba3} using ground-truth geometry and materials. The reference image is rendered using the ground-truth scene lighting, while the remaining columns use the lighting estimated by each respective method. The bottom-left insets show the corresponding direct illumination (shading) maps. AEGIR uses explicit local area emitters, GS-ID combines point lights with an environment map, and IRGS relies solely on environment maps.}
    \label{fig:light_grid_vis}
\end{figure*}

\begin{figure*}[ht] 
    \centering
    
    \newcommand{\IncludegraphicsOrNA}[3]{%
      \ifthenelse{\equal{#2}{NA}}{%
        \fbox{\parbox[c][#3][c]{\dimexpr#1-2\fboxsep-2\fboxrule\relax}{\centering \textbf{N/A}}}%
      }{%
        \includegraphics[width=#1]{#2}%
      }%
    }
    
    \begin{minipage}[t]{0.65\textwidth} 
        \vspace{0pt} 
        \centering
        \captionof{table}{\textbf{Quantitative evaluation of pure lighting estimation.} Metrics are computed by re-rendering each scene in Mitsuba \cite{Mitsuba3} using the light sources estimated by each method and comparing the resulting images against renderings under ground-truth light sources. We report metrics for indoor local lighting and external environment lighting separately.}
        \label{tab:light_grid_quant}
        \vspace{0.5mm}
        
        \resizebox{\columnwidth}{!}{ 
        \begin{tabular}{l c c c c c c}
            \toprule
            \multirow{2}{*}{Method} & \multicolumn{3}{c}{Indoor Lighting} & \multicolumn{3}{c}{Environment Lighting} \\
            \cmidrule(lr){2-4} \cmidrule(lr){5-7}
            & PSNR $\uparrow$ & SSIM $\uparrow$ & LPIPS $\downarrow$  
            & PSNR $\uparrow$ & SSIM $\uparrow$ & LPIPS $\downarrow$ \\
            \midrule
            GS-ID \cite{du2025gsidilluminationdecompositiongaussian} & \underline{28.35} & \underline{0.92} & \underline{0.11} & 26.35 & 0.86 & 0.15 \\
            IRGS \cite{gu2024IRGS}                                   & 11.26 & 0.49 & 0.25 & \underline{28.98} & \underline{0.88} & \underline{0.14} \\
            \midrule
            \textbf{AEGIR (Ours)}                                    & \textbf{31.42} & \textbf{0.94} & \textbf{0.06} & \textbf{30.22} & \textbf{0.92} & \textbf{0.07} \\
            \bottomrule
        \end{tabular}
        }
    \end{minipage}
    \hfill
    \begin{minipage}[t]{0.32\textwidth}
        \vspace{0pt} 
        \centering
        
        \IncludegraphicsOrNA{\linewidth}{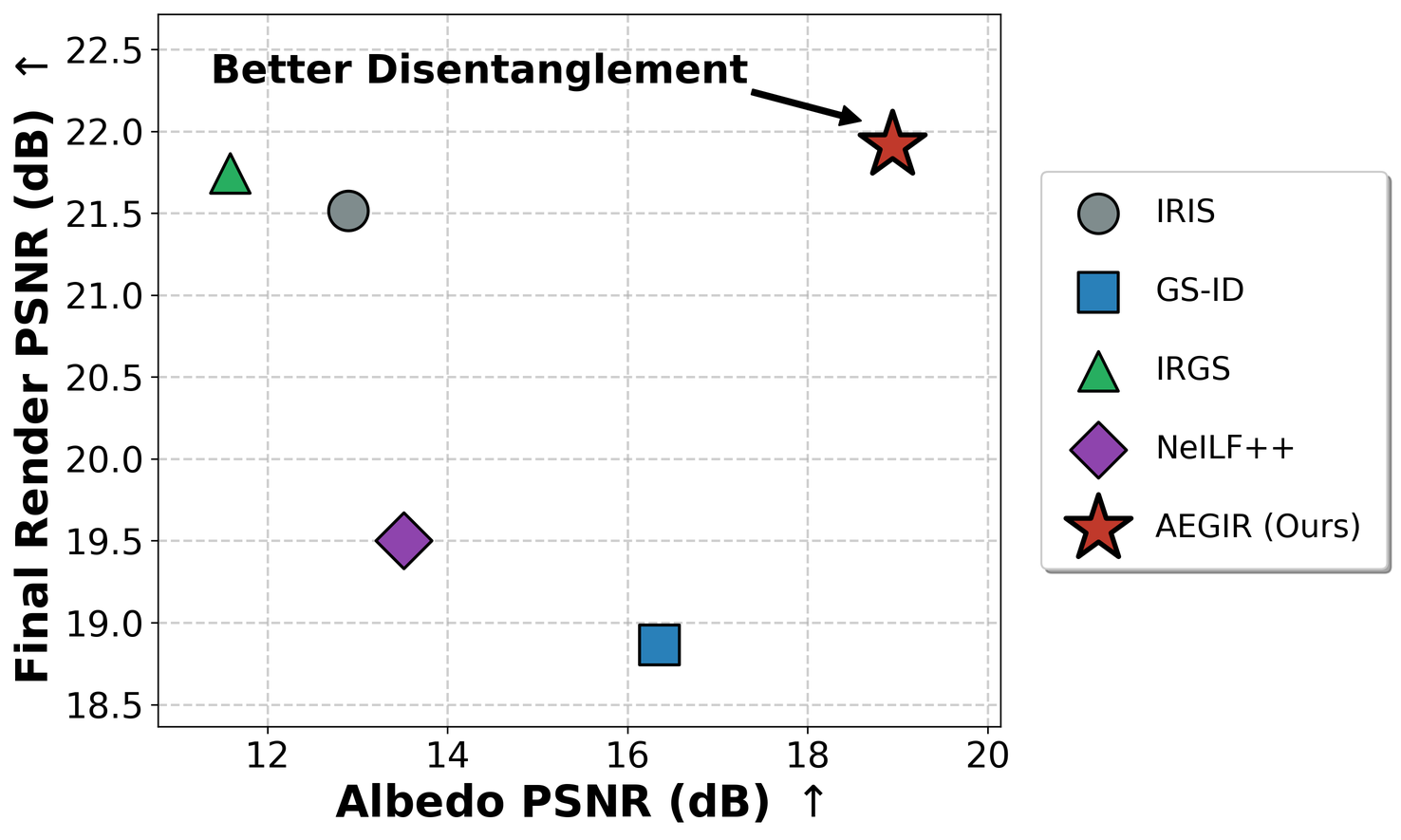}{4.2cm} 
        
        \captionof{figure}{\textbf{Disentanglement plot on synthetic datasets.} Albedo vs. render PSNR shows that AEGIR improves both material accuracy and novel view synthesis.}
        \label{fig:albedo_vs_render_scatter}
    \end{minipage}
    
\end{figure*}

To rigorously evaluate the estimated lighting, we use a controlled Mitsuba \cite{Mitsuba3} pipeline that isolates illumination errors from geometry and materials. Using synthetic scenes from Bitterli's rendering resources \cite{resources16}, we first render ground-truth training images with a simplified BRDF that is consistent with all evaluated methods. After training, we import each method's estimated lighting back into Mitsuba: explicit area emitters for AEGIR, point lights with an environment map for GS-ID, and an environment map alone for IRGS. We then re-render each scene with fixed ground-truth geometry and materials, so the metrics reflect lighting errors only rather than NVS, material, or denoising artifacts. Table \ref{tab:light_grid_quant} and Figure \ref{fig:light_grid_vis} show that the lighting representation strongly affects performance. For indoor lighting, IRGS fails to localize the ceiling light and instead places energy at the windows, causing severe underexposure (11.26 dB PSNR). GS-ID improves localization (28.35 dB) but produces overly concentrated lighting. AEGIR best matches the ground truth (31.42 dB), recovering smoother indoor shading. For environment lighting, IRGS performs better (28.98 dB) since environment maps naturally capture window illumination. GS-ID performs worse (26.35 dB) because its added point lights introduce localized bright spots. AEGIR again performs best (30.22 dB), with smoother window illumination closest to the ground truth. Across both settings, AEGIR achieves the best PSNR, SSIM, and LPIPS. Additional qualitative evaluations of our estimated lighting are provided in Appendix \ref{append:visual}.


\subsection{Other Contributions: Novel View Synthesis and Intrinsic Disentanglement}

\newcommand{\ColorCellHigh}[3]{%
    \pgfmathparse{max(min(100*(#3-#1)/(#2-#1), 100), 0)}%
    \edef\x{\noexpand\cellcolor{red!\pgfmathresult!white}}\x #3%
}
\newcommand{\ColorCellLow}[3]{%
    \pgfmathparse{max(min(100*(#2-#3)/(#2-#1), 100), 0)}%
    \edef\x{\noexpand\cellcolor{red!\pgfmathresult!white}}\x #3%
}

\newcommand{\ColOne}[1]{\ColorCellHigh{19.0}{22.0}{#1}}
\newcommand{\ColTwo}[1]{\ColorCellHigh{0.65}{0.85}{#1}}
\newcommand{\ColThree}[1]{\ColorCellLow{0.10}{0.40}{#1}}

\newcommand{\ColFour}[1]{\ColorCellHigh{11.0}{21.0}{#1}}
\newcommand{\ColFive}[1]{\ColorCellHigh{0.40}{0.80}{#1}}
\newcommand{\ColSix}[1]{\ColorCellLow{0.15}{0.55}{#1}}

\newcommand{\ColSeven}[1]{\ColorCellHigh{18.0}{24.0}{#1}}
\newcommand{\ColEight}[1]{\ColorCellHigh{0.70}{0.90}{#1}}
\newcommand{\ColNine}[1]{\ColorCellLow{0.15}{0.30}{#1}}

\newcommand{\ColTen}[1]{\ColorCellHigh{11.0}{18.0}{#1}}
\newcommand{\ColEleven}[1]{\ColorCellHigh{0.65}{0.85}{#1}}
\newcommand{\ColTwelve}[1]{\ColorCellLow{0.20}{0.35}{#1}}

\newcommand{\ColThirteen}[1]{\ColorCellLow{0.08}{0.17}{#1}}
\newcommand{\ColFourteen}[1]{\ColorCellLow{45.0}{185.0}{#1}}

\begin{table*}[hbt]
    \centering
    \caption{\textbf{Quantitative evaluation on Hypersim and FIPT-Synthetic datasets.} We report the performance of novel view synthesis, intrinsic material decomposition, and the optimization time (in minutes) for each method across both synthetic benchmarks.}
    \vspace{0.5mm} 
    
    \renewcommand{\arraystretch}{1.2} 
    
    \resizebox{\textwidth}{!}{
    \begin{tabular}{l c c c c c c c c c c c c c c}
        \toprule
        
        \multirow{3}{*}{Method} & \multicolumn{6}{c}{Hypersim} & \multicolumn{7}{c}{FIPT-Synthetic} & \multicolumn{1}{c}{\multirow{3}{*}{Time (m)}} \\
        \cmidrule(lr){2-7} \cmidrule(lr){8-14}
        
         & \multicolumn{3}{c}{Novel View Synthesis} & \multicolumn{3}{c}{Albedo} & \multicolumn{3}{c}{Novel View Synthesis} & \multicolumn{3}{c}{Albedo} & \multicolumn{1}{c}{\multirow{2}{*}{\begin{tabular}[c]{@{}c@{}}Roughness\\ MSE $\downarrow$\end{tabular}}} & \multicolumn{1}{c}{} \\
        \cmidrule(lr){2-4} \cmidrule(lr){5-7} \cmidrule(lr){8-10} \cmidrule(lr){11-13}
        
         & \multicolumn{1}{c}{PSNR $\uparrow$} & \multicolumn{1}{c}{SSIM $\uparrow$} & \multicolumn{1}{c}{LPIPS $\downarrow$} & \multicolumn{1}{c}{PSNR $\uparrow$} & \multicolumn{1}{c}{SSIM $\uparrow$} & \multicolumn{1}{c}{LPIPS $\downarrow$} & \multicolumn{1}{c}{PSNR $\uparrow$} & \multicolumn{1}{c}{SSIM $\uparrow$} & \multicolumn{1}{c}{LPIPS $\downarrow$} & \multicolumn{1}{c}{PSNR $\uparrow$} & \multicolumn{1}{c}{SSIM $\uparrow$} & \multicolumn{1}{c}{LPIPS $\downarrow$} & \multicolumn{1}{c}{} & \multicolumn{1}{c}{} \\
        \midrule
        
        IRIS \cite{lin2025iris}            & 20.01 & 0.68 & 0.36 & 11.29 & 0.43 & 0.52 & \textbf{22.89} & \textbf{0.87} & \textbf{0.18} & 14.51 & \underline{0.77} & \underline{0.23} & \underline{0.096} & 60.0 \\
        GS-ID \cite{du2025gsidilluminationdecompositiongaussian} & 19.47 & 0.75 & 0.25 & \underline{17.67} & \underline{0.73} & 0.31 & 18.26 & 0.78 & 0.21 & \underline{15.04} & 0.76 & 0.26 & 0.124 & \underline{55.0} \\
        IRGS \cite{gu2024IRGS}                                   & \underline{21.22} & \underline{0.80} & 0.23 & 11.78 & 0.66 & 0.37 & 22.27 & 0.82 & 0.20 & 11.39 & 0.67 & 0.31 & 0.149 & \textbf{50.0} \\
        NeILF++ \cite{zhang2023neilf++}                          & 20.95 & 0.79 & \underline{0.15} & 14.95 & 0.66 & \underline{0.28} & 18.05 & 0.73 & 0.26 & 12.08 & 0.70 & 0.29 & 0.163 & 180.0 \\
        \midrule
        \textbf{AEGIR (Ours)}                                    & \textbf{21.50} & \textbf{0.84} & \textbf{0.12} & \textbf{20.55} & \textbf{0.78} & \textbf{0.19} & \underline{22.83} & \underline{0.86} & \underline{0.19} & \textbf{17.33} & \textbf{0.80} & \textbf{0.21} & \textbf{0.092} & \underline{55.0} \\
        \bottomrule
    \end{tabular}
    }
    \label{tab:synthetic_nvs}
\end{table*}

\newcommand{\ColU}[1]{\ColorCellHigh{14.0}{19.0}{#1}}
\newcommand{\ColV}[1]{\ColorCellHigh{0.78}{0.85}{#1}}
\newcommand{\ColW}[1]{\ColorCellLow{0.18}{0.40}{#1}}

\newcommand{\ColE}[1]{\ColorCellHigh{13.0}{18.5}{#1}}
\newcommand{\ColR}[1]{\ColorCellHigh{0.60}{0.80}{#1}}
\newcommand{\ColT}[1]{\ColorCellLow{0.20}{0.36}{#1}}

\newcommand{\ColF}[1]{\ColorCellHigh{20.0}{22.5}{#1}}
\newcommand{\ColG}[1]{\ColorCellHigh{0.80}{0.86}{#1}}
\newcommand{\ColH}[1]{\ColorCellLow{0.17}{0.28}{#1}}

\newcommand{\ColJ}[1]{\ColorCellLow{45.0}{185.0}{#1}}

\begin{table*}[ht]
    \centering
    \caption{\textbf{Quantitative evaluation on real-world datasets.} We report the performance of novel view synthesis and optimization times (in minutes) across the ScanNet++, Replica, and FIPT-Real benchmarks.}
    \vspace{0.5mm} 
    
    \renewcommand{\arraystretch}{1.2} 
    
    \resizebox{\textwidth}{!}{
    \begin{tabular}{l c c c c c c c c c c}
        \toprule
        
        \multirow{2}{*}{Method} & \multicolumn{3}{c}{ScanNet++} & \multicolumn{3}{c}{Replica} & \multicolumn{3}{c}{FIPT-Real} & \multicolumn{1}{c}{\multirow{2}{*}{Time (m)}} \\
        \cmidrule(lr){2-4} \cmidrule(lr){5-7} \cmidrule(lr){8-10}
        
         & \multicolumn{1}{c}{PSNR $\uparrow$} & \multicolumn{1}{c}{SSIM $\uparrow$} & \multicolumn{1}{c}{LPIPS $\downarrow$} & \multicolumn{1}{c}{PSNR $\uparrow$} & \multicolumn{1}{c}{SSIM $\uparrow$} & \multicolumn{1}{c}{LPIPS $\downarrow$} & \multicolumn{1}{c}{PSNR $\uparrow$} & \multicolumn{1}{c}{SSIM $\uparrow$} & \multicolumn{1}{c}{LPIPS $\downarrow$} & \multicolumn{1}{c}{} \\
        \midrule
        
        IRIS \cite{lin2025iris}            & 15.22 & 0.80 & 0.23 & 17.41 & 0.75 & 0.27 & \underline{21.96} & \underline{0.84} & \textbf{0.19} & 60.0 \\
        GS-ID \cite{du2025gsidilluminationdecompositiongaussian} & \underline{18.02} & \underline{0.82} & \underline{0.21} & 17.05 & 0.73 & 0.29 & 20.31 & 0.82 & \underline{0.21} & \underline{55.0} \\
        IRGS \cite{gu2024IRGS}                                   & 17.48 & \underline{0.82} & 0.39 & 13.64 & 0.63 & 0.34 & 21.23 & 0.81 & 0.26 & \textbf{50.0} \\
        NeILF++ \cite{zhang2023neilf++}                          & 17.54 & \textbf{0.83} & 0.22 & \underline{17.79} & \underline{0.77} & \underline{0.25} & 20.67 & 0.82 & \underline{0.21} & 180.0 \\
        \midrule
        \textbf{AEGIR (Ours)}                                    & \textbf{18.55} & \textbf{0.83} & \textbf{0.20} & \textbf{21.34} & \textbf{0.84} & \textbf{0.22} & \textbf{24.98} & \textbf{0.87} & \textbf{0.19} & \underline{55.0} \\
        \bottomrule
    \end{tabular}
    }
    \label{tab:real_world_nvs}
\end{table*}

\begin{figure*}[ht]
    \centering
    
    \newcommand{\gridimage}[1]{%
        \parbox[c][0.06\textwidth][c]{0.08\textwidth}{%
            \centering%
            \IfFileExists{#1}{%
                \includegraphics[width=0.08\textwidth, height=0.06\textwidth, keepaspectratio]{#1}%
            }{%
                \small\textbf{N/A}%
            }%
        }%
    }

    \newcommand{\quadblock}[4]{%
        \setlength{\tabcolsep}{0pt}
        \renewcommand{\arraystretch}{0}
        \setlength{\arrayrulewidth}{0.5pt}
        
        \begin{tabular}[c]{|c|c|} 
            \hline
            \gridimage{#1} & \gridimage{#2} \\
            \hline
            \gridimage{#3} & \gridimage{#4} \\
            \hline
        \end{tabular}%
    }
    
    \setlength{\tabcolsep}{1pt} 
    \renewcommand{\arraystretch}{1.2} 
    
    \begin{tabular}{cccccc}
        \small Reference & \small \textbf{AEGIR (Ours)} & \small GS-ID \cite{du2025gsidilluminationdecompositiongaussian} & \small IRGS \cite{gu2024IRGS} & \small IRIS \cite{lin2025iris} & \small NeILF++ \cite{zhang2023neilf++} \\
        
        \quadblock{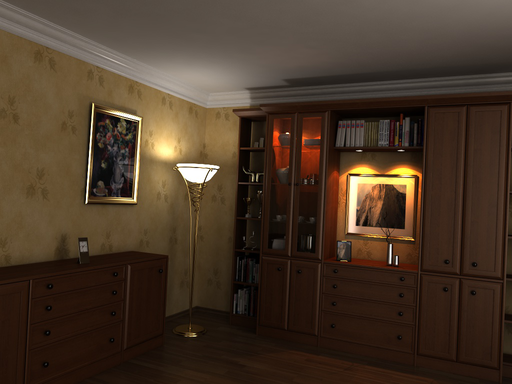}{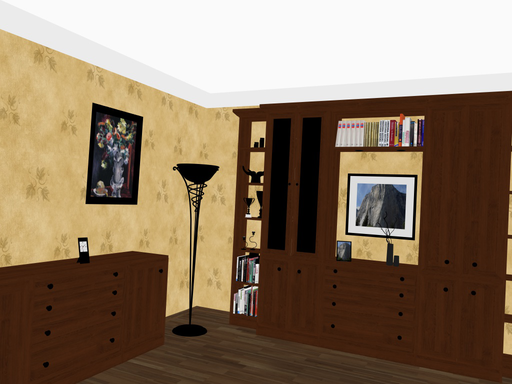}{figures/nvs/ai_001_004/rough.png}{figures/nvs/ai_001_004/metal.png} &
        
        \quadblock{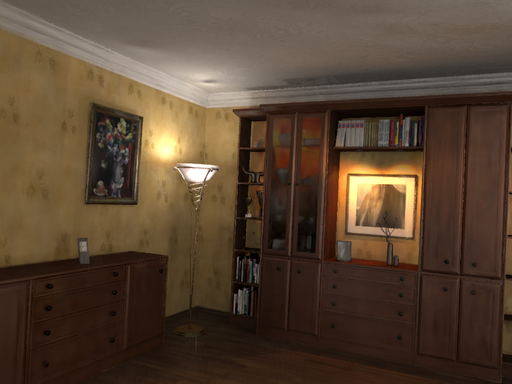}{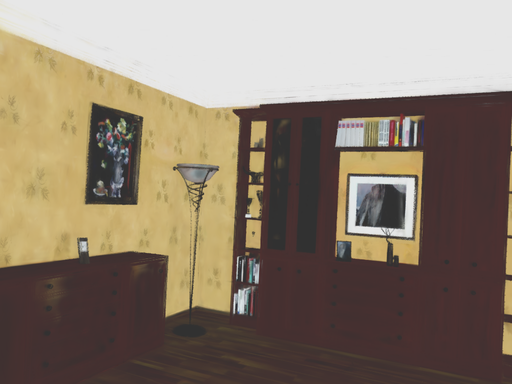}{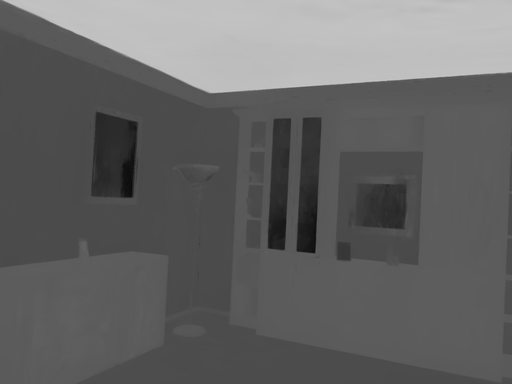}{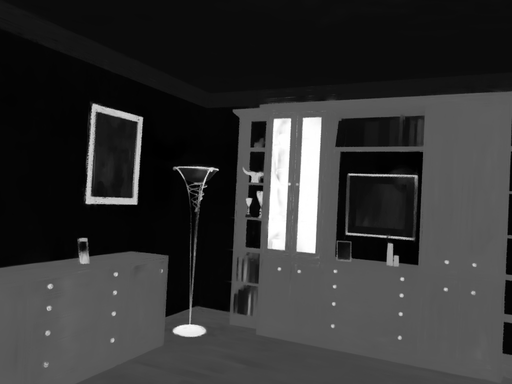} &
        
        \quadblock{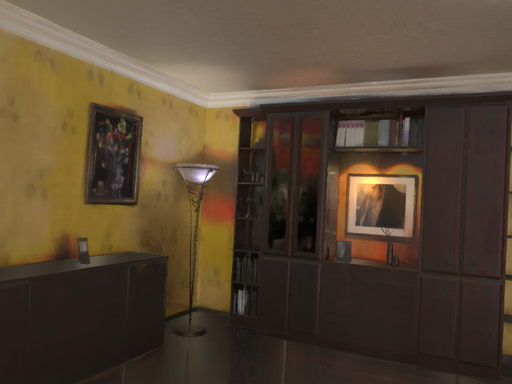}{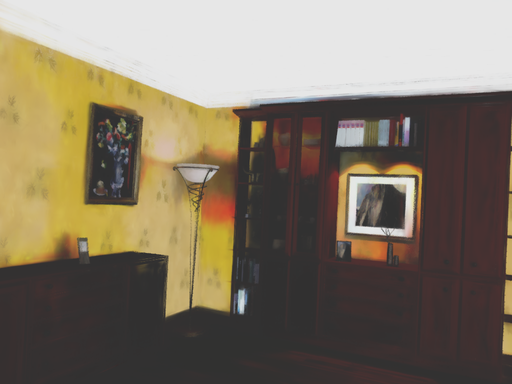}{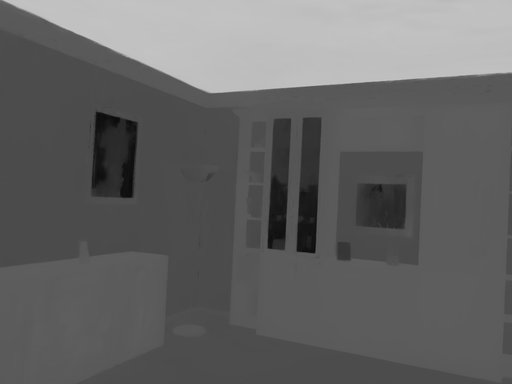}{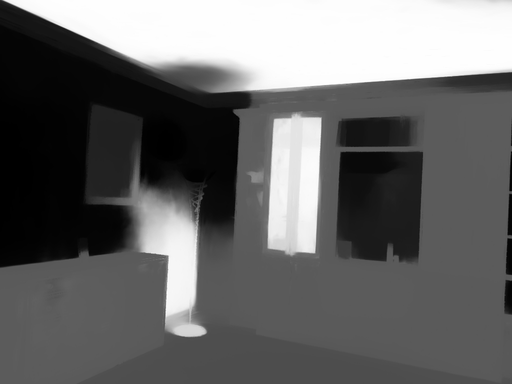} &
        
        \quadblock{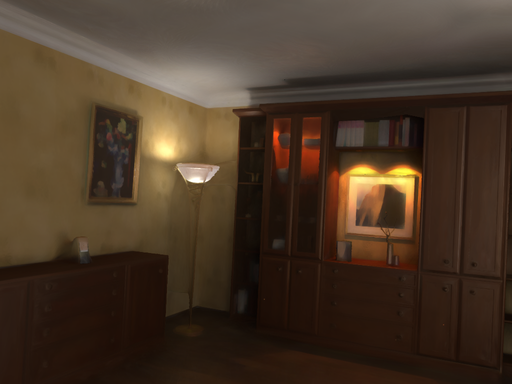}{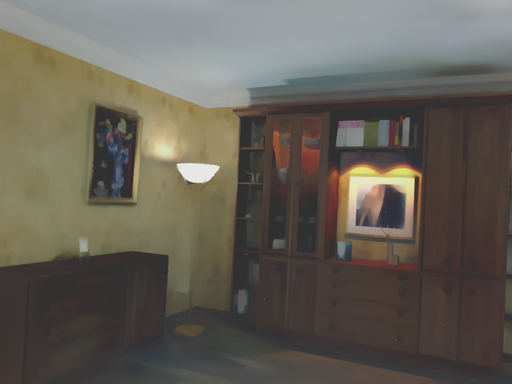}{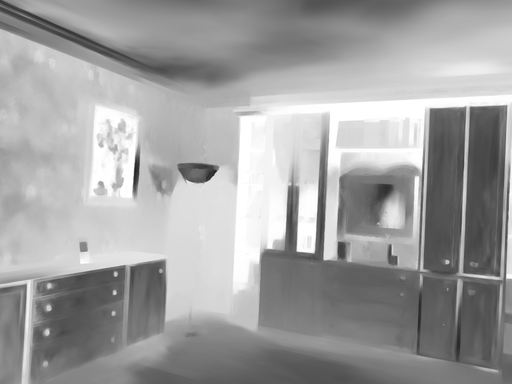}{figures/nvs/ai_001_004/irgs/metal.png} &
        
        \quadblock{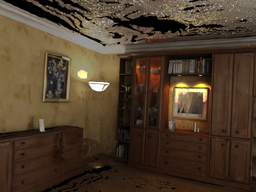}{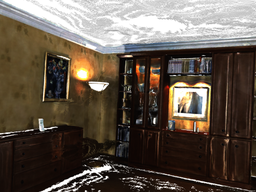}{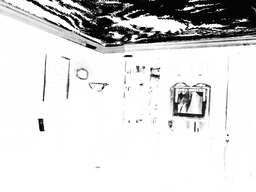}{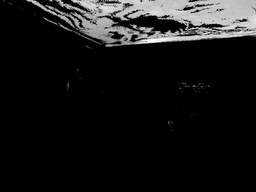} &
        
        \quadblock{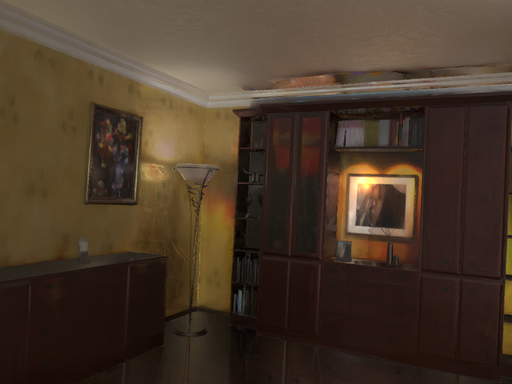}{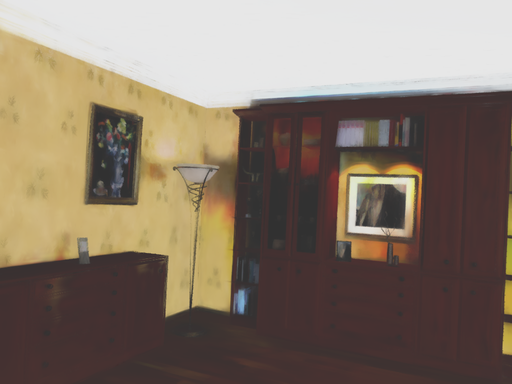}{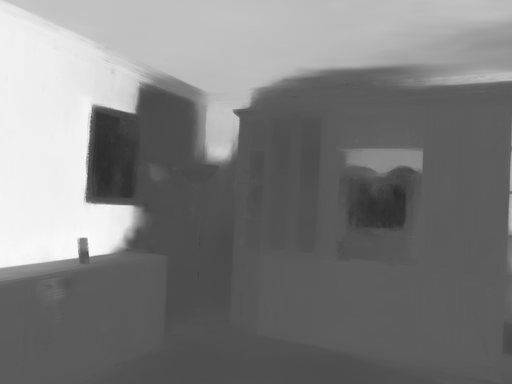}{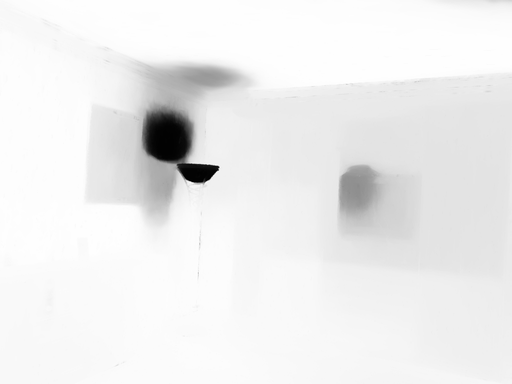} \\[7mm]

        \quadblock{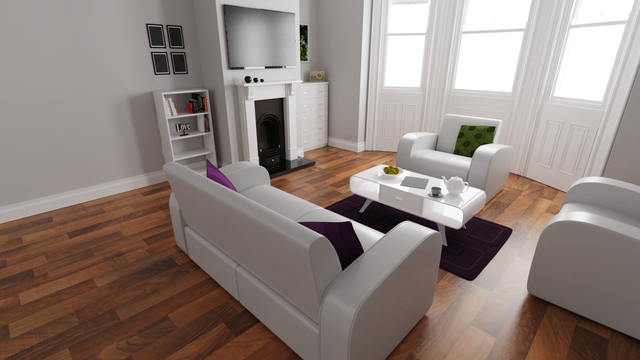}{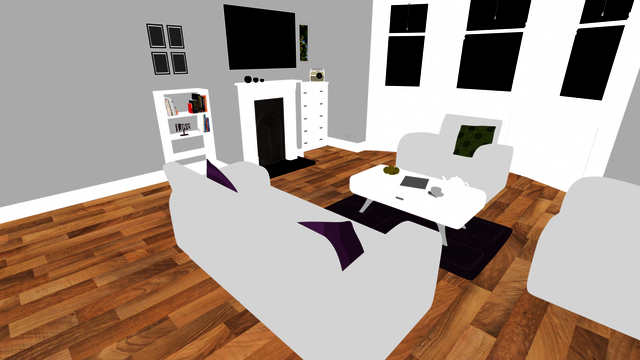}{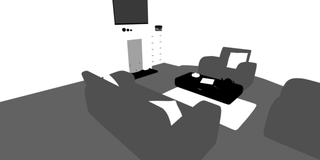}{figures/nvs/livingroom/metal.png} &
        
        \quadblock{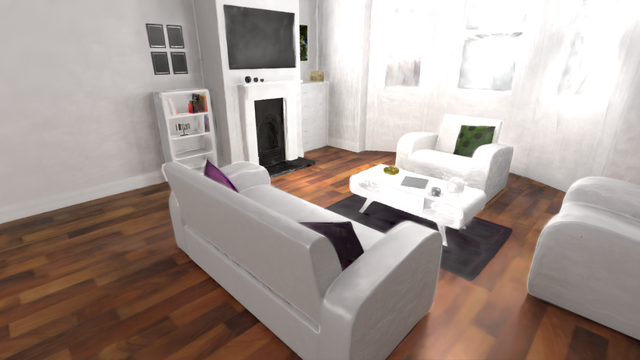}{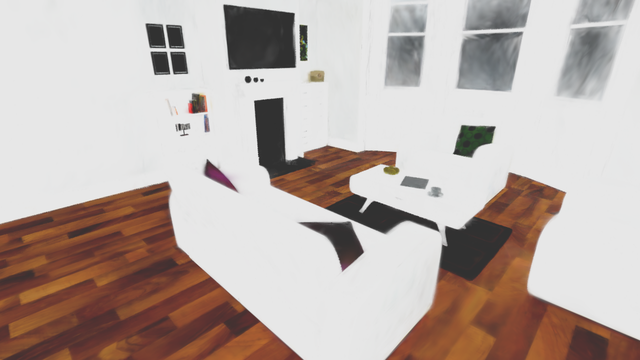}{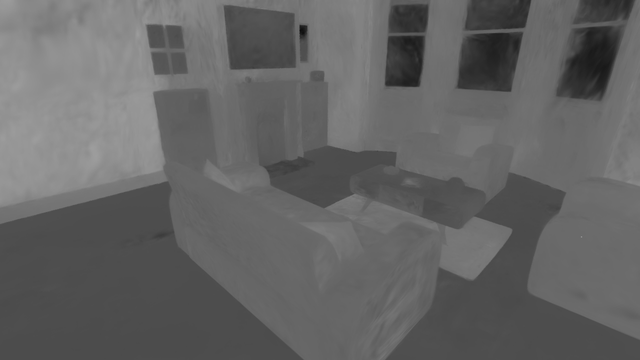}{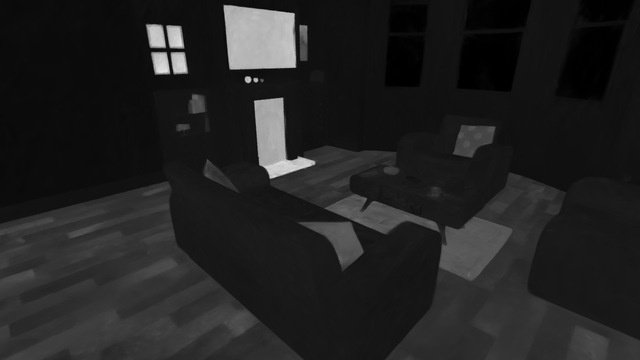} &
        
        \quadblock{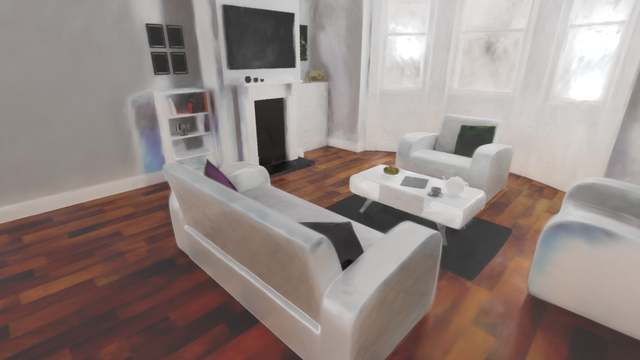}{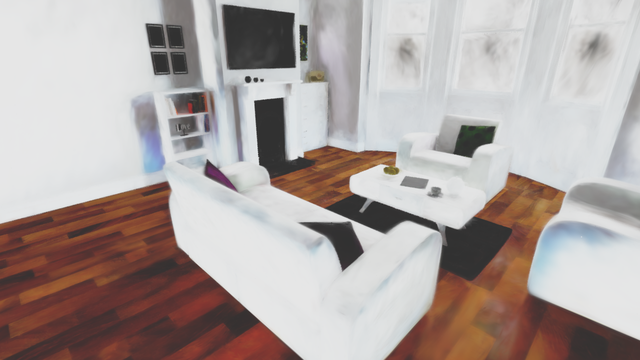}{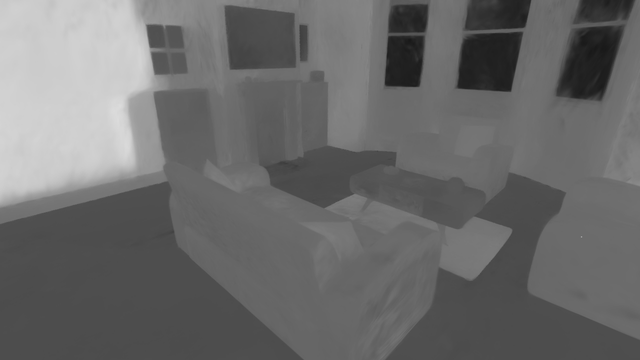}{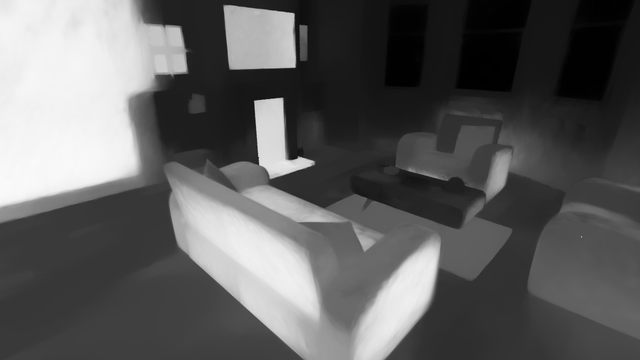} &
        
        \quadblock{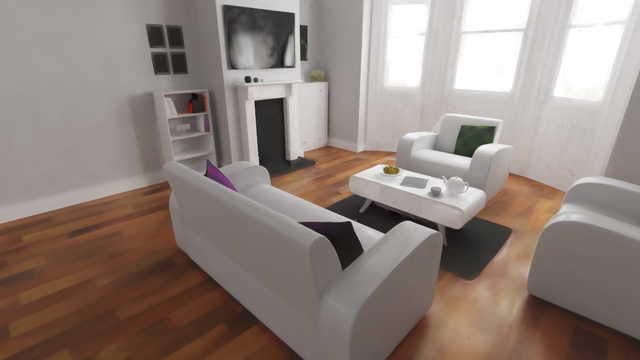}{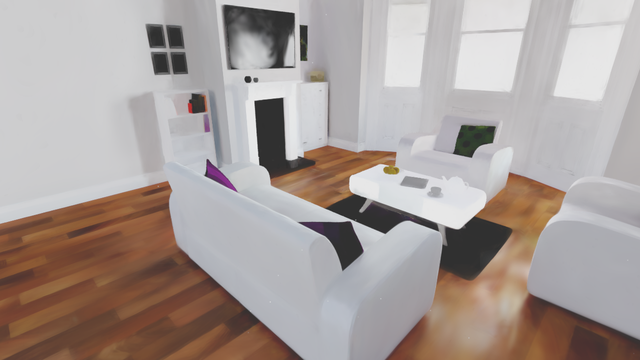}{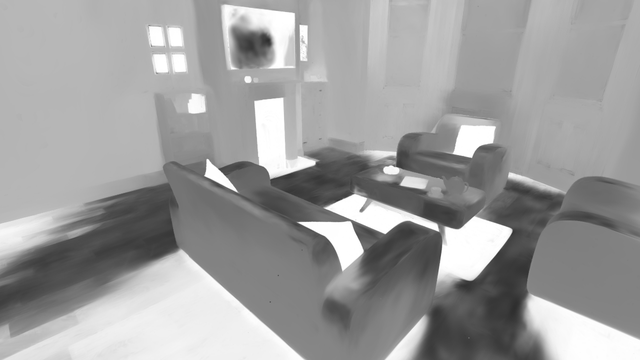}{figures/nvs/livingroom/irgs/metal.png} &
        
        \quadblock{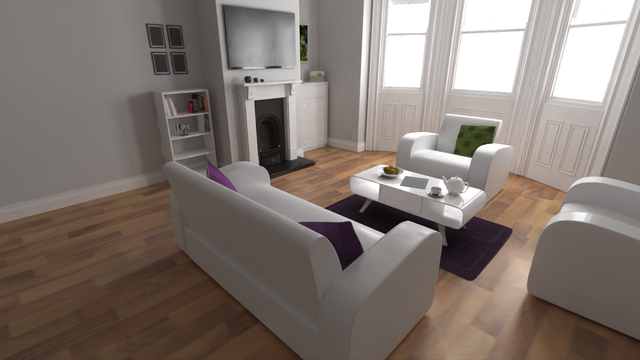}{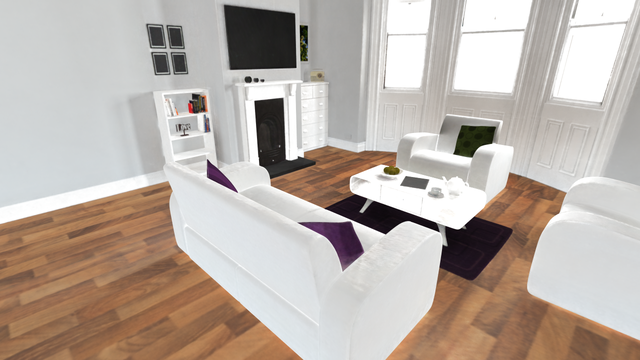}{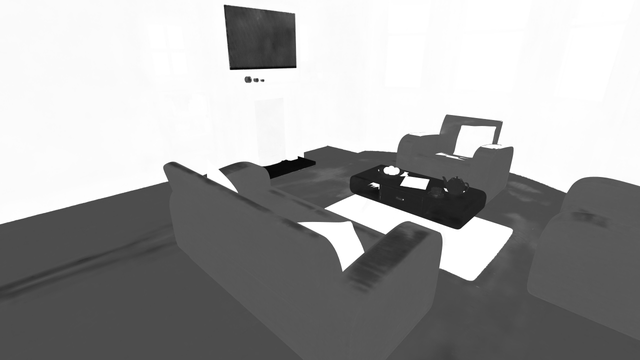}{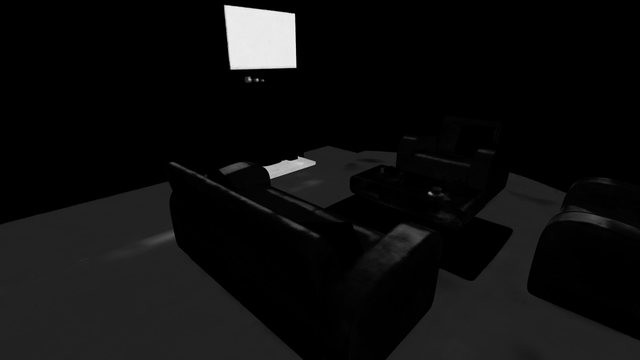} &
        
        \quadblock{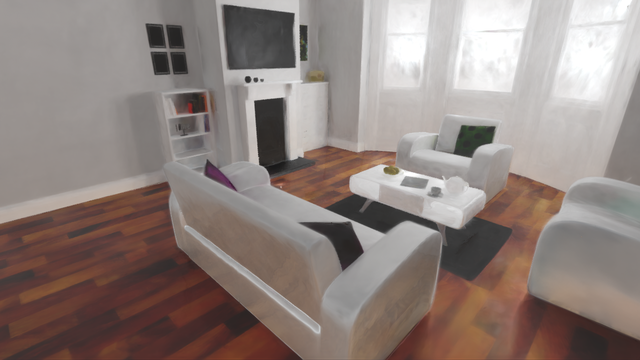}{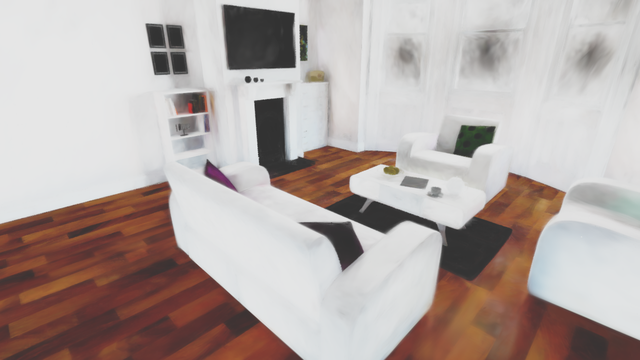}{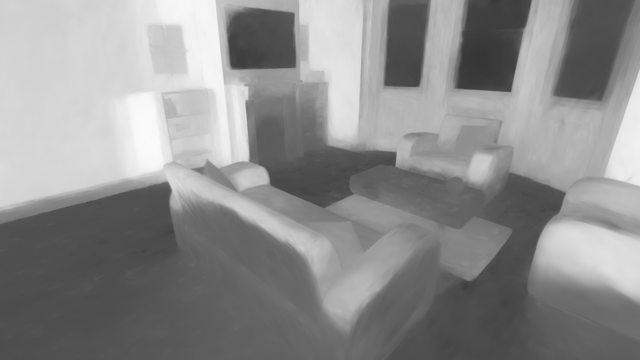}{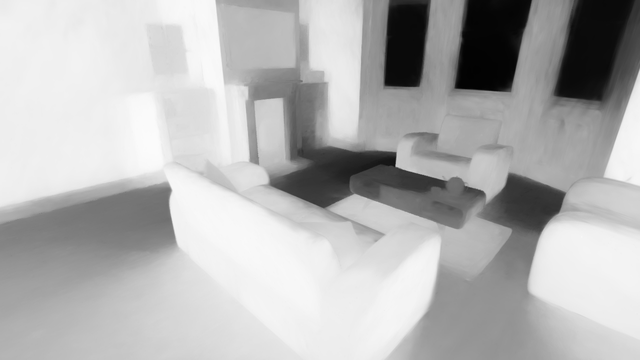} \\[7mm]

        \quadblock{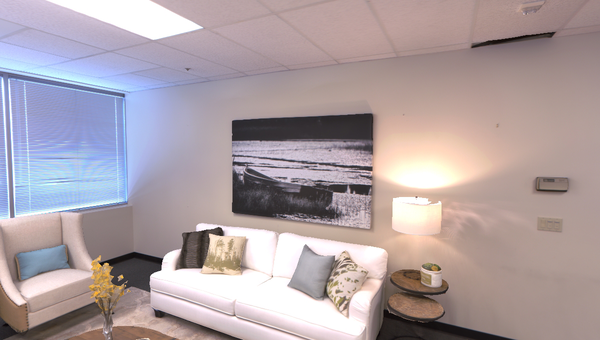}{figures/nvs/room0/albedo.png}{figures/nvs/room0/rough.png}{figures/nvs/room0/metal.png} &
        
        \quadblock{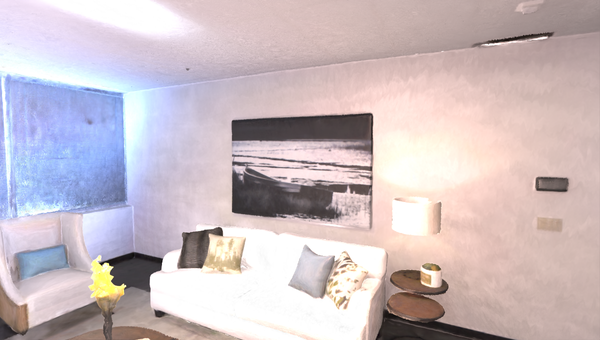}{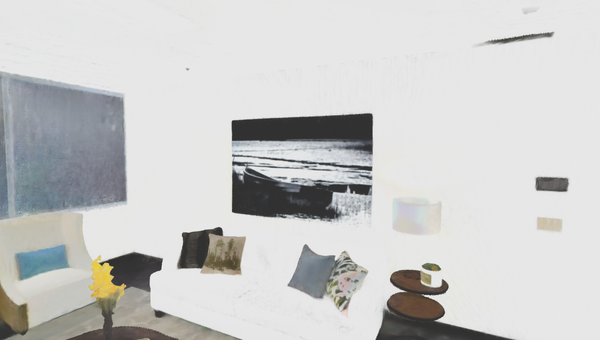}{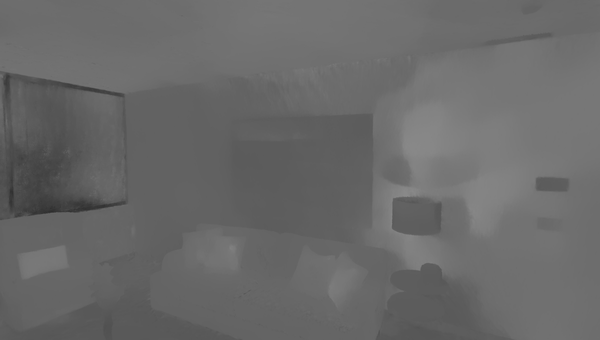}{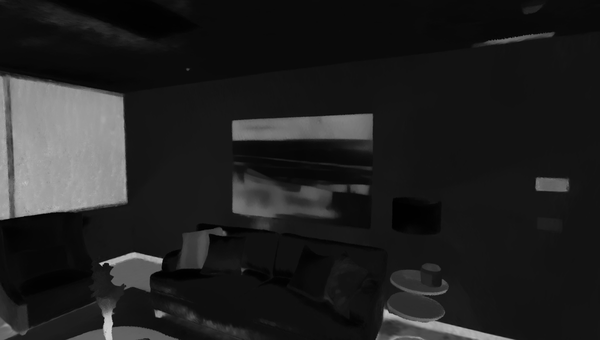} &
        
        \quadblock{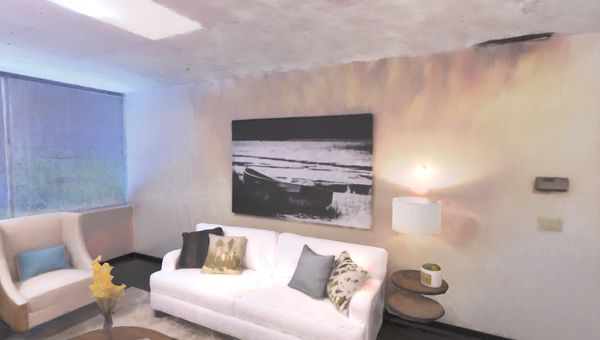}{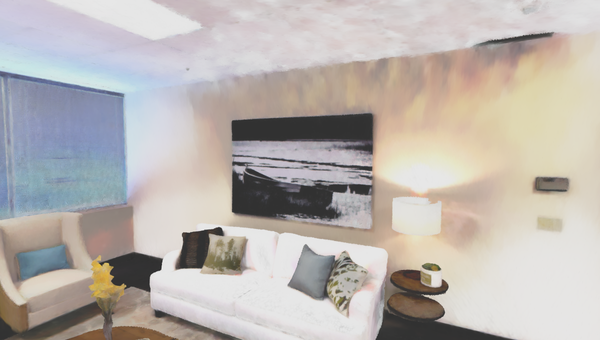}{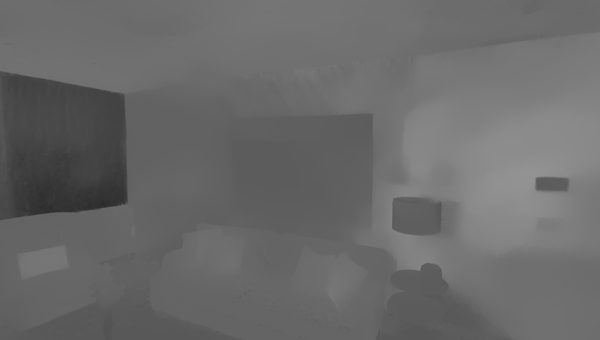}{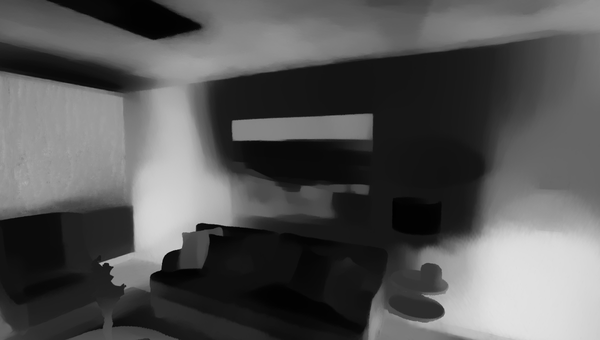} &
        
        \quadblock{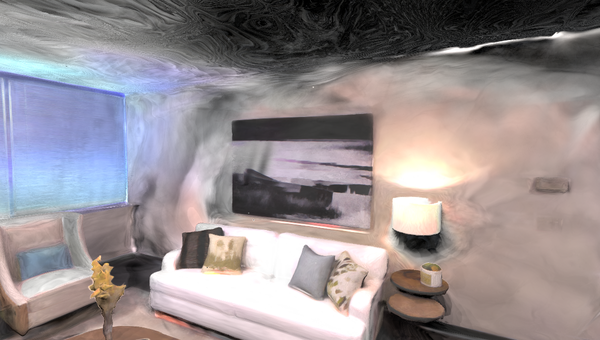}{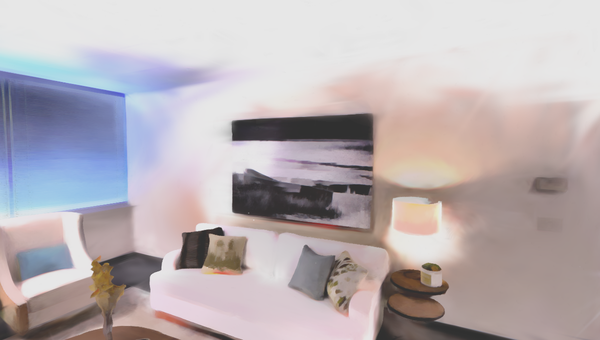}{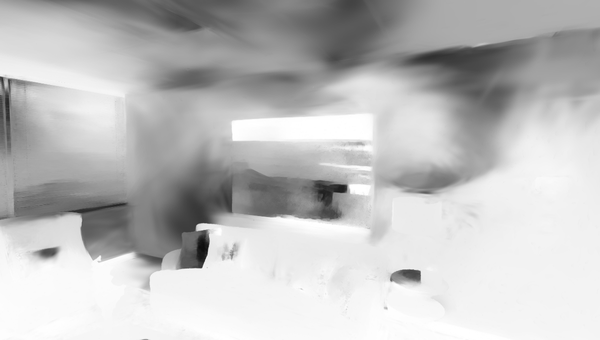}{figures/nvs/room0/irgs/metal.png} &
        
        \quadblock{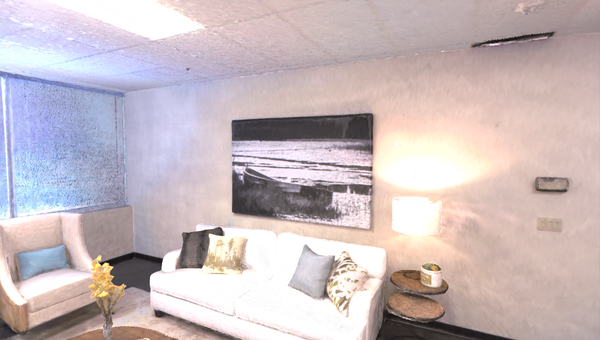}{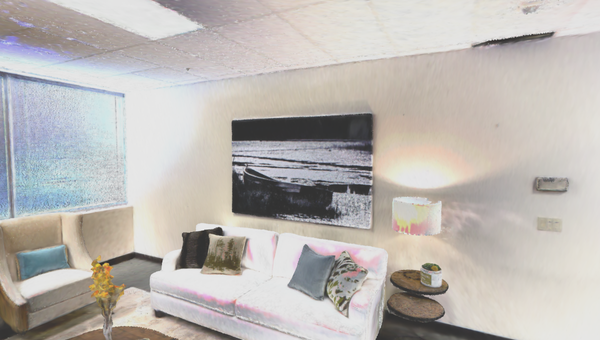}{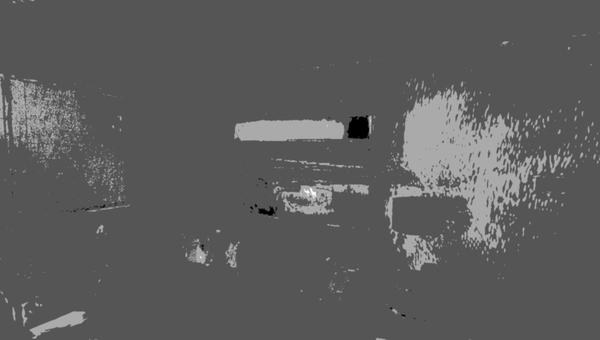}{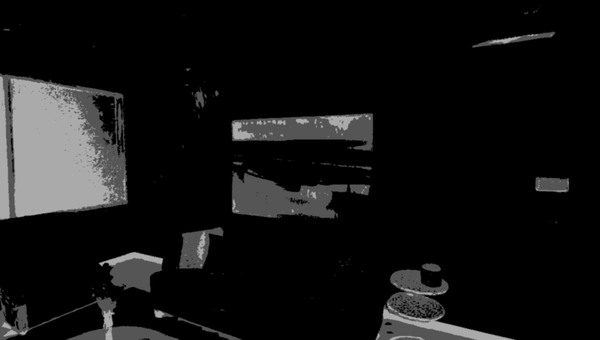} &
        
        \quadblock{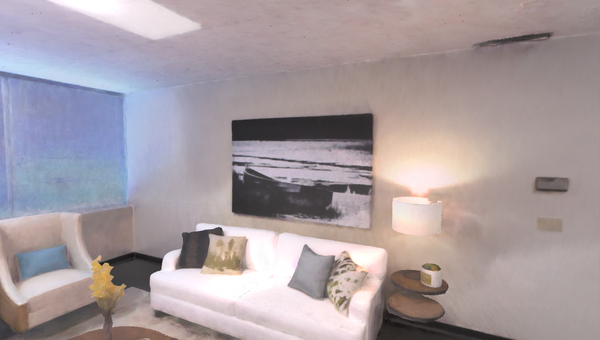}{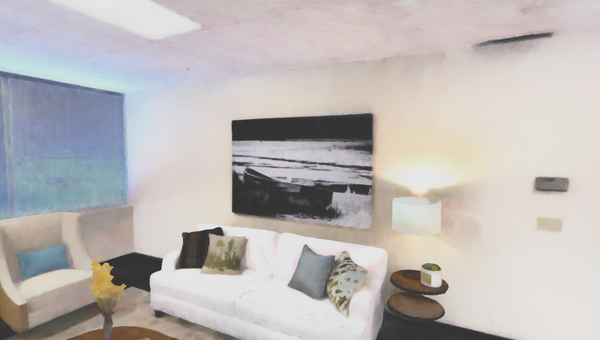}{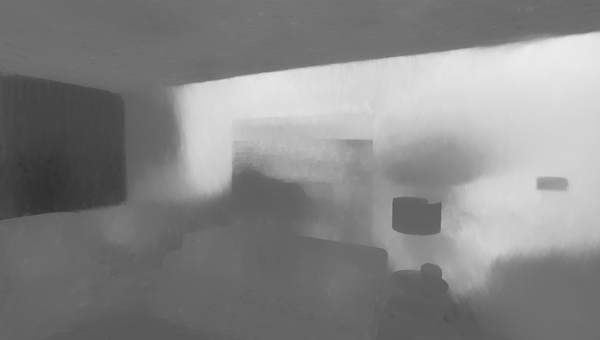}{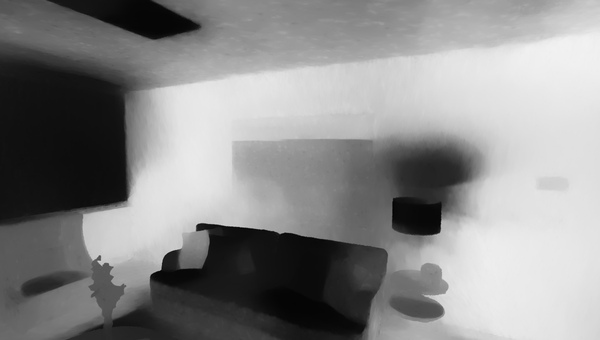} \\
    \end{tabular}
    \vspace{0.5mm}
    \caption{\textbf{Qualitative comparison of novel view synthesis and intrinsic decomposition.} Each block shows a $2 \times 2$ grid of the final render (top-left), albedo (top-right), roughness (bottom-left), and metallic (bottom-right) maps. ``N/A'' in the reference column indicates missing annotations in the dataset, while ``N/A'' in the results indicates outputs not produced by the corresponding method.}
    \label{fig:intrinsic_quad_layout}
\end{figure*}

As shown in Table \ref{tab:real_world_nvs}, AEGIR achieves the best novel view synthesis (NVS) performance across ScanNet++, Replica, and FIPT-Real, matching or leading the best score in PSNR, SSIM, and LPIPS. Our method remains efficient with a total optimization time of 55 minutes (including geometry initialization), making it approximately three times faster than NeILF++ and comparable to other baselines. On synthetic datasets (Table \ref{tab:synthetic_nvs}), our method achieves the best overall performance on the Hypersim benchmark in both NVS and albedo recovery, outperforming the next best method by nearly 3 dB in albedo PSNR. On FIPT-Synthetic, our method remains competitive with IRIS, which achieves slightly higher NVS scores on this dataset by baking residual illumination into the surface albedo. In contrast, our approach maintains better material recovery without relying on such artifacts. This trade-off is further illustrated in Figure \ref{fig:albedo_vs_render_scatter}, which shows that baseline methods often trade material accuracy for rendering quality, while our method balances both. These quantitative results are supported by the visual comparisons in Figure \ref{fig:intrinsic_quad_layout}. Our pipeline produces sharper renderings and cleaner intrinsic maps, avoiding the blur and baked-in illumination artifacts common in other approaches. This strong performance comes from effective physical disentanglement. Explicit geometric area emitters reduce ambiguity between lighting and materials, while emitter surface sampling and differentiable visibility ensure accurate light transport. Combined with diffusion priors and material--lighting regularization, this produces a stable, physically plausible decomposition. As shown in Figure \ref{fig:applications_grid}, this high-quality factorization also enables downstream applications such as virtual object insertion and controlled relighting through direct manipulation of individual geometric light sources.


\subsection{Ablation Study}

\begin{table}[ht]
    \centering
    \caption{\textbf{Quantitative ablation study.} NVS metrics are averaged over Replica and FIPT-Synthetic, while albedo metrics are measured on FIPT-Synthetic.}
    \label{tab:ablation_study}
    \vspace{0.5mm} 
    
    \resizebox{0.75\linewidth}{!}{%
        \begin{tabular}{l c c c c c c}
            \toprule
            \multirow{2}{*}{Configuration} & \multicolumn{3}{c}{\begin{tabular}[c]{@{}c@{}}NVS\\(Replica + FIPT-Synthetic avg.)\end{tabular}} & \multicolumn{3}{c}{\begin{tabular}[c]{@{}c@{}}Albedo\\(FIPT-Synthetic)\end{tabular}} \\
            \cmidrule(lr){2-4} \cmidrule(lr){5-7}
             & \multicolumn{1}{c}{PSNR $\uparrow$} & \multicolumn{1}{c}{SSIM $\uparrow$} & \multicolumn{1}{c}{LPIPS $\downarrow$} & \multicolumn{1}{c}{PSNR $\uparrow$} & \multicolumn{1}{c}{SSIM $\uparrow$} & \multicolumn{1}{c}{LPIPS $\downarrow$} \\
            \midrule
            W/o Focus Map ($W_f$)        & 17.47 & 0.74 & 0.24 & 17.25 & 0.78 & 0.23 \\
            W/o Diffusion Priors ($\mathcal{L}_{prior}$)      & 19.59 & 0.80 & \underline{0.21} & 15.11 & 0.74 & 0.26 \\
            W/o Material TV ($\mathcal{L}_{TV}$)            & 19.05 & 0.78 & 0.23 & 15.96 & 0.75 & 0.25 \\
            W/o Light Regularization ($\mathcal{L}_{light}$)   & 18.19 & 0.76 & 0.23 & 17.24 & \underline{0.79} & 0.23 \\
            W/o Adaptive Control       & 17.83 & 0.74 & 0.24 & 17.21 & 0.77 & 0.24 \\
            W/o Emitter Initialization & \underline{20.18} & \underline{0.82} & \underline{0.21} & \underline{17.30} & \underline{0.79} & \underline{0.22} \\
            \midrule
            \textbf{AEGIR (Full Method)} & \textbf{22.08} & \textbf{0.85} & \textbf{0.20} & \textbf{17.33} & \textbf{0.80} & \textbf{0.21} \\
            \bottomrule
        \end{tabular}%
    }
\end{table}

\begin{figure*}[ht]
    \centering
    
    \newcommand{\safeimg}[2]{%
        \parbox[c]{#1}{%
            \centering
            \IfFileExists{#2}{%
                \includegraphics[width=\linewidth]{#2}%
            }{%
                \setlength{\fboxsep}{0pt}%
                \setlength{\fboxrule}{0.4pt}%
                \framebox{\parbox[c][1cm][c]{\dimexpr\linewidth-0.8pt\relax}{\centering \fontsize{5}{6}\selectfont \textbf{N/A}}}%
            }%
        }%
    }

    \newcommand{\ablationcell}[3]{%
        \parbox[t]{0.123\linewidth}{%
            \centering
            \safeimg{\linewidth}{#1}\\[0.3mm]
            \safeimg{0.48\linewidth}{#2}\hfill%
            \safeimg{0.48\linewidth}{#3}%
        }%
    }

    \newcommand{\gtcell}[1]{%
        \parbox[t]{0.123\linewidth}{%
            \centering
            \safeimg{\linewidth}{#1}\\[1mm]
            \phantom{\safeimg{0.48\linewidth}{#1}}%
        }%
    }
    
    \newcommand{\IncludegraphicsOrNA}[3]{%
      \ifthenelse{\equal{#2}{NA}}{%
        \fbox{\parbox[c][#3][c]{\dimexpr#1-2\fboxsep-2\fboxrule\relax}{\centering \textbf{N/A}}}%
      }{%
        \includegraphics[width=#1]{#2}%
      }%
    }

    \setlength{\tabcolsep}{1pt} 
    \begin{tabular}{@{} *{8}{c} @{}}
        \scriptsize Reference 
        & \scriptsize \textbf{Full Method} 
        & \scriptsize \shortstack{W/o Focus Map} 
        & \scriptsize \shortstack{W/o Diff. Prior} 
        & \scriptsize \shortstack{W/o Mat. TV} 
        & \scriptsize \shortstack{W/o Light Reg.} 
        & \scriptsize \shortstack{W/o Ada. Control} 
        & \scriptsize \shortstack{W/o Emitter Init.} \\[1.5mm]
        
        \gtcell{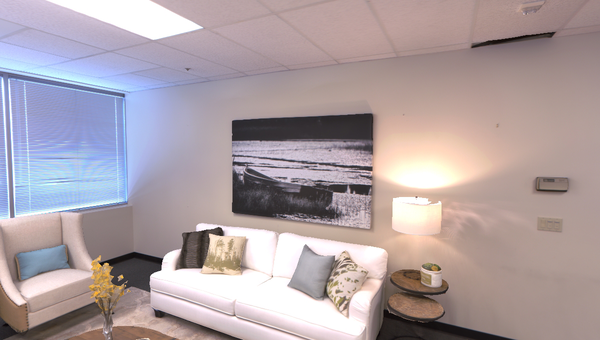} & 
        \ablationcell{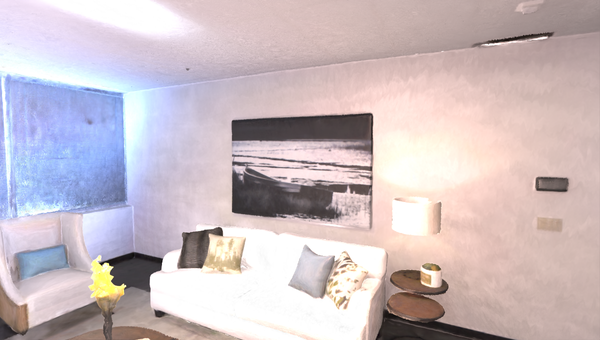}{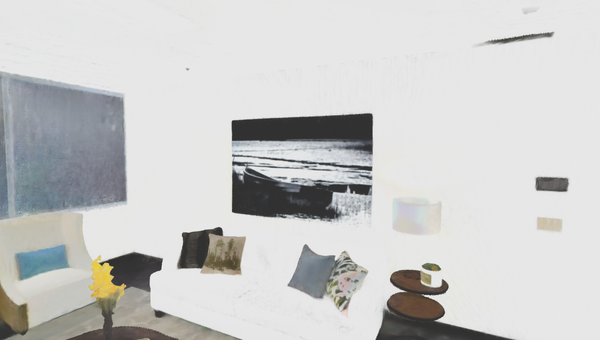}{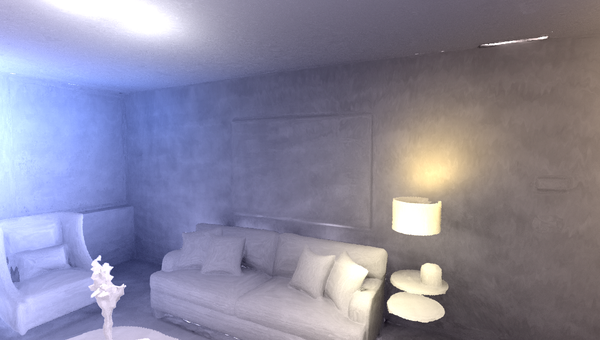} & 
        \ablationcell{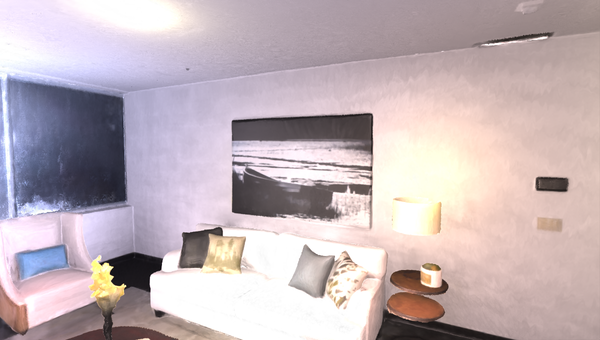}{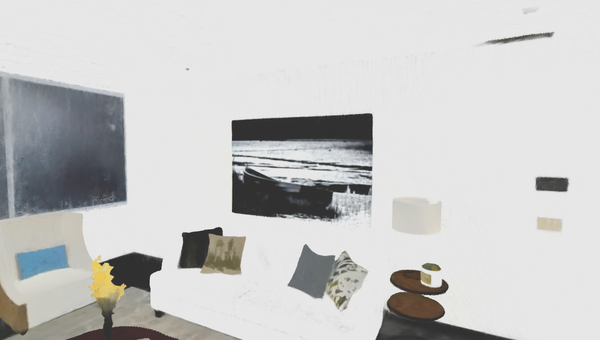}{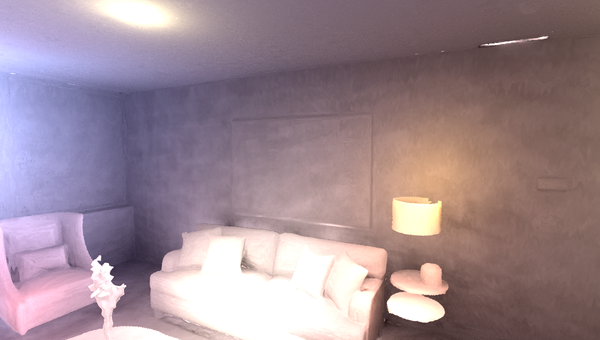} & 
        \ablationcell{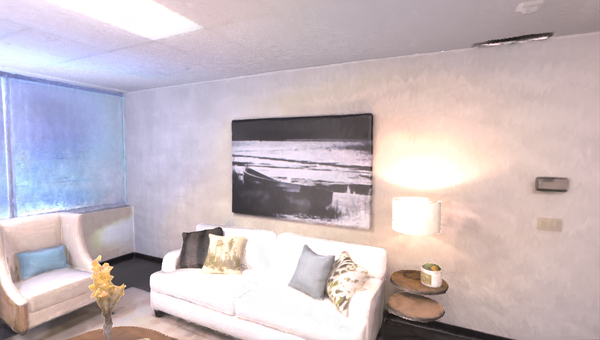}{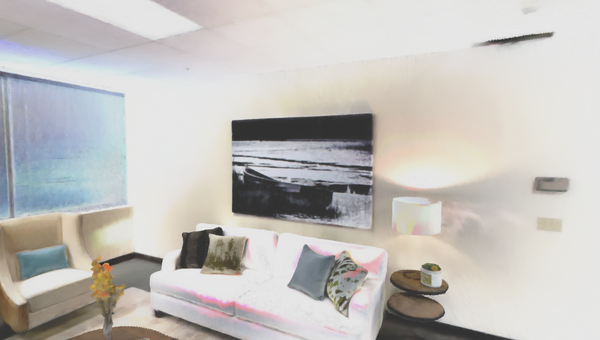}{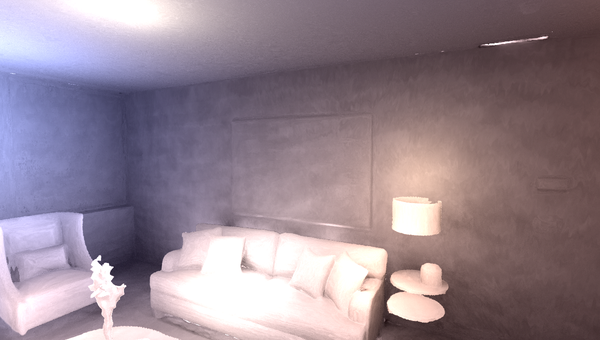} & 
        \ablationcell{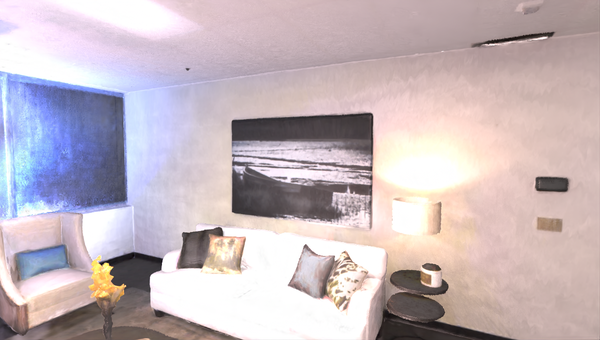}{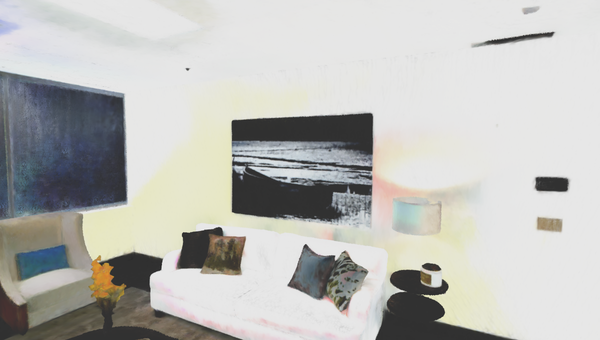}{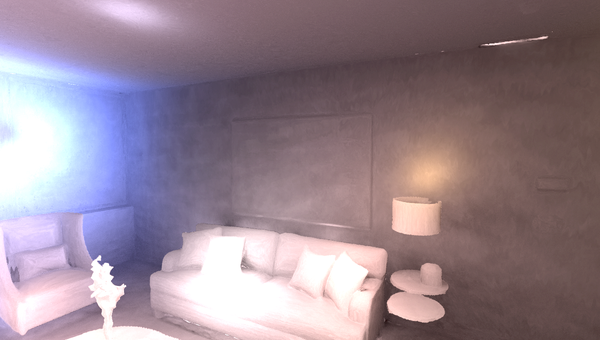} & 
        \ablationcell{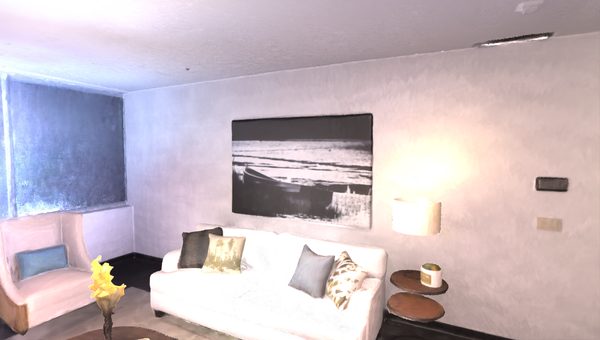}{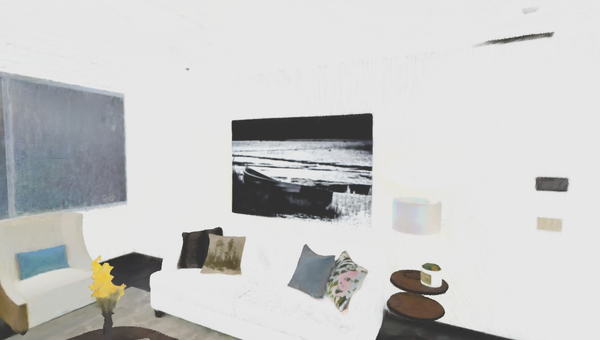}{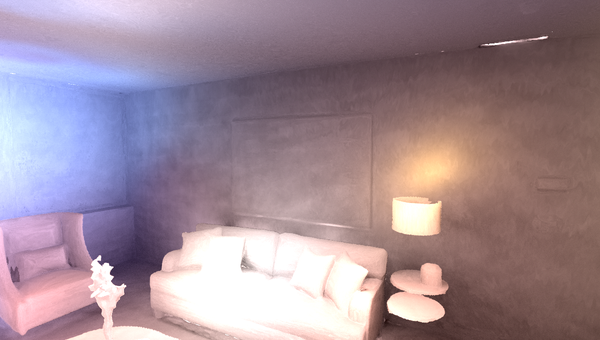} & 
        \ablationcell{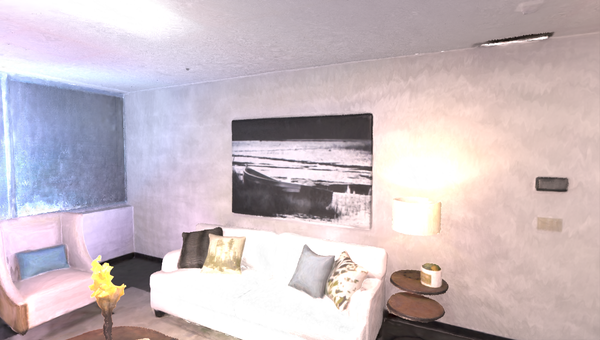}{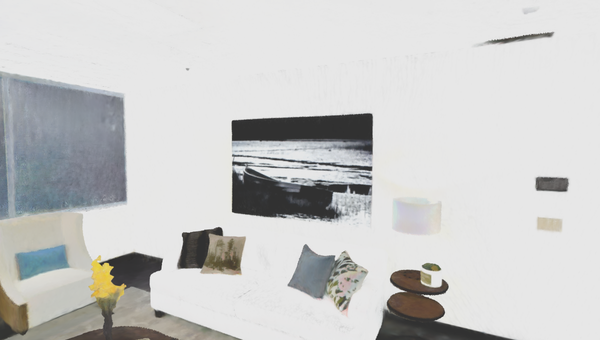}{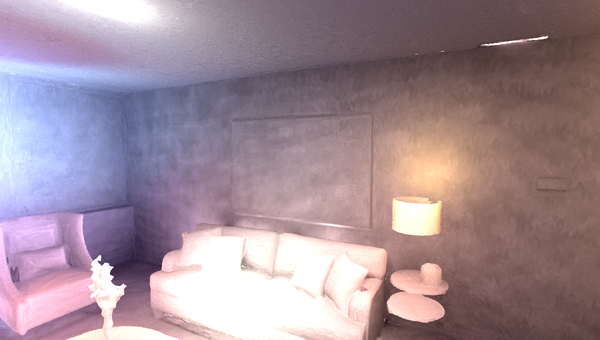} & 
        \ablationcell{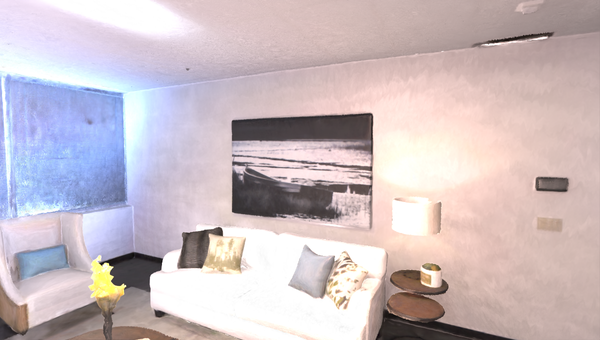}{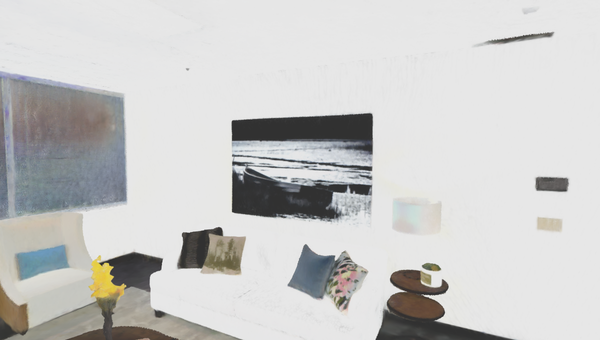}{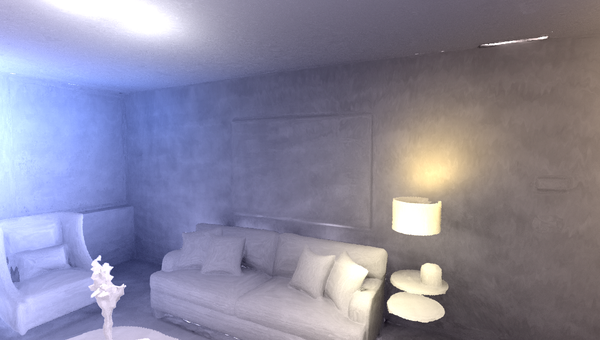} \\
    \end{tabular}
    
    \vspace{0mm}
    \caption{\textbf{Qualitative ablation study.} The top row shows novel view renders; the bottom row shows estimated albedo (left) and direct illumination (right). Removing key components embeds illumination artifacts in the albedo and degrades illumination estimation.}
    \label{fig:ablation_qualitative}

    \vspace{0.5mm}

    \renewcommand{\arraystretch}{0.5} 
    \begin{tabular}{@{}c@{\hspace{1mm}}c@{\hspace{1mm}}c@{\hspace{2mm}}c@{\hspace{1mm}}c@{\hspace{1mm}}c@{}}
        {\scriptsize Render} & {\scriptsize Relight} & {\scriptsize Insertion} & {\scriptsize Render} & {\scriptsize Relight} & {\scriptsize Insertion} \\[1mm]
        \IncludegraphicsOrNA{0.162\linewidth}{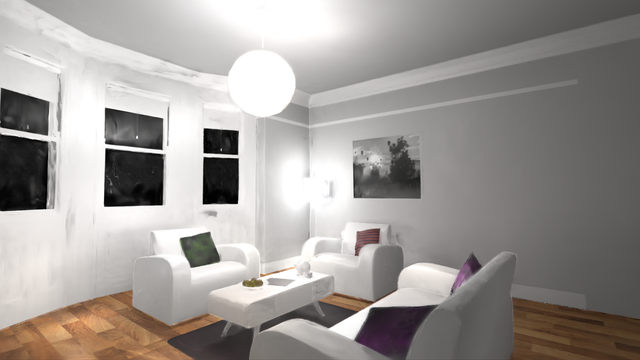}{1.5cm} & 
        \IncludegraphicsOrNA{0.162\linewidth}{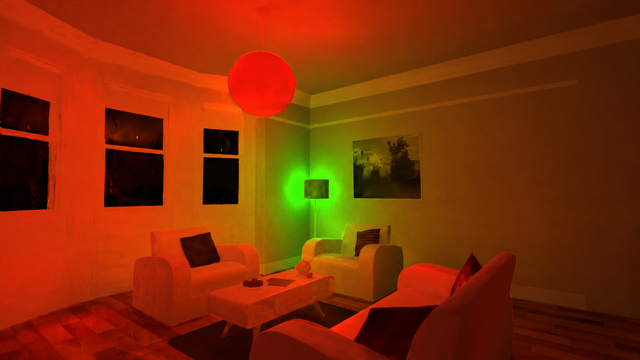}{1.5cm} & 
        \IncludegraphicsOrNA{0.162\linewidth}{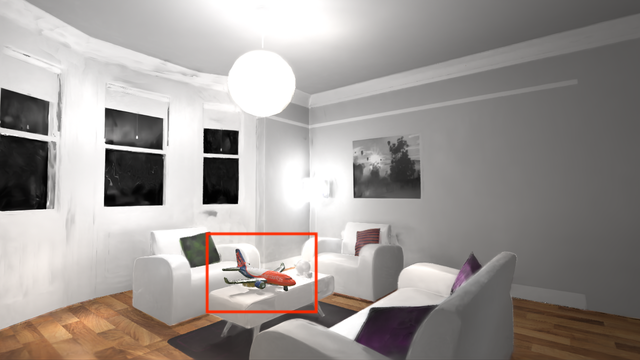}{1.5cm} & 
        \IncludegraphicsOrNA{0.162\linewidth}{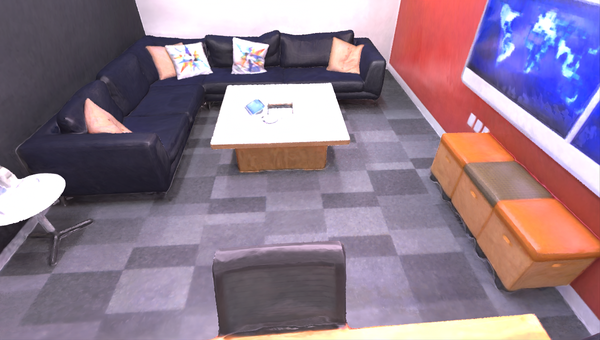}{1.5cm} & 
        \IncludegraphicsOrNA{0.162\linewidth}{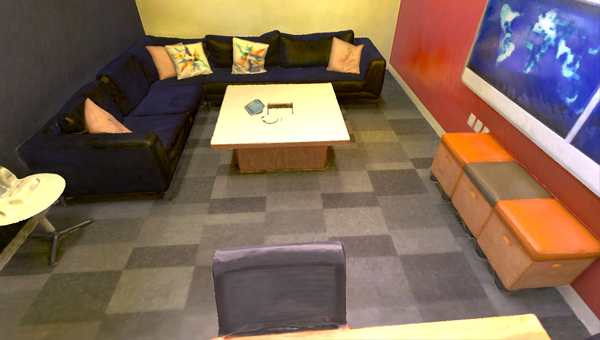}{1.5cm} & 
        \IncludegraphicsOrNA{0.162\linewidth}{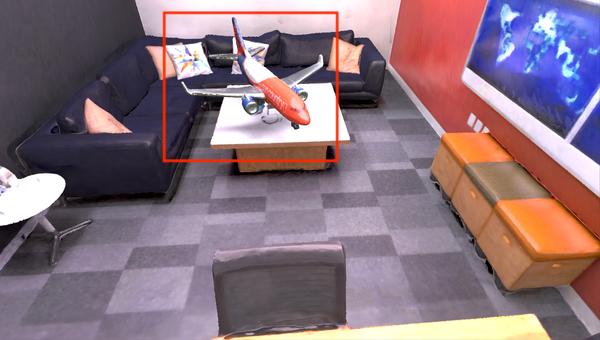}{1.5cm} \\
    \end{tabular}
    
    \vspace{0mm}
    \caption{\textbf{Downstream applications.} By factorizing illumination into explicit area emitters, our framework supports controlled scene relighting and virtual object insertion.}
    \label{fig:applications_grid}

\end{figure*}

We evaluate the contribution of each component to novel view synthesis and albedo accuracy (Table \ref{tab:ablation_study}), as well as qualitative illumination estimation and intrinsic disentanglement (Figure \ref{fig:ablation_qualitative}) in scenes with complex indoor lighting from the Replica and FIPT-Synthetic datasets. Removing the focus multiplier weakens the detection of localized lighting patterns. Meanwhile, disabling diffusion priors causes a severe drop in the quantitative albedo metrics, as illumination effects become incorrectly embedded into the estimated albedo. Omitting the material TV loss leads to noisy textures due to inconsistencies in DiffusionRenderer outputs. For illumination estimation, removing light regularization or adaptive control results in degenerate solutions, where extra light sources are introduced and area light sizes are overestimated. Ablating emitter initialization causes the smallest degradation among the lighting-control ablations, especially for albedo, but still reduces NVS quality and slows convergence. This indicates its role as a useful photometric prior and demonstrates that our method can accurately estimate illumination from shading cues only, without requiring light sources to be directly visible in the input images. As supported by the quantitative metrics in Table \ref{tab:ablation_study} and the visual results in Figure \ref{fig:ablation_qualitative}, the full AEGIR model achieves the best balance between rendering quality, albedo accuracy, and stable illumination estimation, demonstrating the necessity of each component.

\section{Conclusion and Future Work}\label{sec:conclusion}
We present AEGIR, an inverse rendering framework that models explicit local area emitters for physically grounded illumination estimation. By modeling the spatial extent, shape, and anisotropic emission of real-world light sources within a 2D Gaussian Splatting representation, AEGIR bridges neural scene modeling with physically based rendering. Combined with differentiable soft shadow tracing, multiple importance sampling, and regularization for materials and lighting, our approach reduces inverse rendering ambiguity and prevents illumination from being baked into albedo. This results in a more consistent decomposition of geometry, materials, and lighting, with strong performance in material recovery, illumination estimation, and novel view synthesis.

Finally, several technical limitations remain. First, shading artifacts can arise from noisy normals. Second, the cost of shadow tracing increases with the number of emitters, motivating more efficient sampling methods such as ReSTIR \cite{10.1145/3386569.3392481}. Finally, our current material model does not handle mirror-like or transparent materials. More broadly, photorealistic manipulation of real environments carries risks of misuse in the generation of deceptive media, highlighting the need for robust methods to detect synthetic media. Despite these challenges, AEGIR provides a reliable approach for recovering illumination under complex lighting conditions. By explicitly factorizing illumination together with material properties, it enables flexible and realistic scene editing.

\medskip
\bibliographystyle{unsrt}
\bibliography{ref}

\appendix

\section{Deferred Rendering Details}\label{append:render}
This section provides a detailed mathematical description of the deferred rendering pipeline of AEGIR, including the microfacet BRDF, light transport integration, differentiable visibility, and final pixel evaluation.

\subsection{Microfacet BRDF Model}

To model the interaction between light and surfaces, AEGIR uses a standard microfacet BRDF. The reflectance $f_r(\mathbf{v}, \omega)$ for view direction $\mathbf{v}$ and light direction $\omega$ is decomposed as $f_r=f_d+f_s$ into diffuse and specular lobes. The diffuse component is:
\begin{equation}
f_d = \frac{\mathbf{c} (1 - m)}{\pi}
\label{eq:app_brdf_diffuse}
\end{equation}
where $\mathbf{c}$ is albedo and $m$ is metallic.
The specular component follows the GGX model:
\begin{equation}
f_s = \frac{D(\mathbf{h}, \alpha_r) \cdot G(\mathbf{v}, \omega, \alpha_r) \cdot F(\mathbf{v}, \mathbf{h})}{4 (\mathbf{n} \cdot \mathbf{v})_+(\mathbf{n} \cdot \omega)_+ + \epsilon}
\label{eq:app_brdf_specular}
\end{equation}
where $\mathbf{h}$ is the half-vector and $\alpha_r = r^2$ is derived from the surface roughness $r$. $D$ is the GGX normal distribution, $G$ is the Smith visibility term, and $F$ is computed using Schlick's approximation. The base reflectance is defined as $F_0 = 0.04 (1 - m) + \mathbf{c} m$.

\subsection{Monte Carlo Sampling and MIS Approximation}

AEGIR estimates the outgoing radiance $L_o(x, \mathbf{v})$ at surface point $x$ using two sampling strategies: cosine-weighted BRDF sampling and explicit emitter sampling. Let $N_{\mathrm{brdf}}$ be the number of BRDF samples. For emitter sampling, we allocate the same number of samples to each of the $N_e$ active emitters. We use $K=\lfloor N_{\mathrm{light}}/N_e \rfloor$ samples per emitter, giving $N'_{\mathrm{light}}=N_eK$ light samples and $N=N_{\mathrm{brdf}}+N'_{\mathrm{light}}$ total samples. The Monte Carlo estimate is:
\begin{equation}
L_o(x, \mathbf{v}) \approx
\frac{1}{N}
\sum_{j=1}^{N}
\frac{
f_r(\mathbf{v}, \omega_j)
L_i(x, \omega_j)
(\mathbf{n} \cdot \omega_j)_+
}{
\tilde{\mu}_{\mathrm{mix}}(\omega_j, x)
}.
\label{eq:app_monte_carlo}
\end{equation}

For BRDF samples drawn from the cosine-weighted hemisphere, the solid-angle PDF is:
\begin{equation}
\mu_{\mathrm{brdf}}(\omega) =
\frac{(\mathbf{n} \cdot \omega)_+}{\pi},
\label{eq:app_pdf_brdf}
\end{equation}
where $(a)_+=\max(a,0)$.

For explicit emitter samples, we define the emitter frame as $\mathbf{R}_e=[\mathbf{u}_e\ \mathbf{v}_e\ \mathbf{w}_e]$. A point on emitter $e$ is sampled by stretching a unit-sphere sample $\boldsymbol{\zeta}$ by the emitter scales $\mathbf{S}_e$, then rotating and translating it into the scene:
\begin{equation}
q_e(\boldsymbol{\zeta}) =
p_e + \mathbf{R}_e
\left(
\mathbf{S}_e \odot \boldsymbol{\zeta}
\right),
\qquad
\|\boldsymbol{\zeta}\|_2 = 1.
\label{eq:app_light_surface_sample}
\end{equation}

This gives a sampled point $q_e$ on the ellipsoidal emitter surface. For a light sample, we first sample a point $q_j$ on an emitter and then form the direction $\omega_j=(q_j-x)/\|q_j-x\|_2$. For a BRDF sample, we first sample the direction $\omega_j$ and then intersect the ray with the explicit emitters. If the ray reaches emitter $e_j$ at point $q_j$, we evaluate the light sampling surrogate. Otherwise, the light sampling term is set to zero:
\begin{equation}
\tilde{\mu}_{\mathrm{light}}(\omega_j, x) =
\begin{cases}
\displaystyle
\frac{1}{N_e}
\frac{\|q_j - x\|_2^2}{A_{e_j} + \epsilon},
& \text{if } \omega_j \text{ reaches emitter } e_j \text{ at } q_j, \\[1.2em]
0,
& \text{otherwise},
\end{cases}
\label{eq:app_pdf_light}
\end{equation}
with the effective area proxy:
\begin{equation}
A_{e_j} =
4\pi
\left(
S_{e_j,u}S_{e_j,v}S_{e_j,w}
\right)^{2/3}.
\label{eq:app_area_proxy}
\end{equation}

The factor $1/N_e$ accounts for the uniform mixture over active emitters. The term $\tilde{\mu}_{\mathrm{light}}$ is not an exact normalized solid-angle PDF, but a stable surrogate for MIS weighting. Distance attenuation is captured through the actual emitter point $q_j$ and its distance to the shaded point $x$.

The combined MIS weighting term is:
\begin{equation}
\tilde{\mu}_{\mathrm{mix}}(\omega_j, x) =
\frac{
N_{\mathrm{brdf}} \mu_{\mathrm{brdf}}(\omega_j)
+
N'_{\mathrm{light}} \tilde{\mu}_{\mathrm{light}}(\omega_j, x)
}{
N_{\mathrm{brdf}} + N'_{\mathrm{light}}
}.
\label{eq:app_mis_balance}
\end{equation}

\subsection{Differentiable Shadow Tracing and Illumination}

To determine the visibility of area emitters, AEGIR traces shadow rays through the 2D Gaussian geometry. Visibility is evaluated as a soft and differentiable mask by comparing the distance to the light against the depth of the intersected geometry. Note that the sigmoid in Equation \ref{eq:visibility_term} depends on the scale of the scene. In practice, we normalize scenes before optimization and use fixed visibility parameters in normalized coordinates, but a scene-dependent steepness parameter is an equivalent implementation choice.

The total incident light $L_i(x, \omega_i)$ is computed as the visibility-weighted sum in Equation \ref{eq:combined_light}. We explicitly restrict this term to direct emitter lighting plus a single secondary bounce, rather than attempting full multi-bounce global illumination. During training, the secondary-bounce term $L_i^{bounce}(x,\omega_i)$ is approximated by tracing to the first secondary Gaussian hit and querying its color features, represented as Spherical Harmonics (SH), following the efficient proxy used in IRGS \cite{gu2024IRGS}. This SH query provides a cheap approximation for occluded or residual indirect illumination while the optimization learns the explicit emitters and materials.

At inference time, we disable the SH proxy. Instead, for the same single-bounce term, we sample a diffuse hemisphere direction at the secondary hit, trace it through the 2D Gaussian scene, and evaluate the recovered area-emitter radiance if the ray reaches an explicit emitter. If the secondary ray does not reach an emitter, the bounce contribution is set to zero. Thus, final rendering uses the optimized emitter representation for both direct lighting and the cheap one-bounce indirect estimate, with SH used only during training for stability. Subsequently, the total incident light $L_i(x, \omega_i)$ is aggregated using the Monte Carlo integration defined in Equation \ref{eq:app_monte_carlo} to compute the outgoing high dynamic range (HDR) radiance $L_o(x, \mathbf{v})$ at the surface point. This linear HDR radiance is then mapped to low dynamic range (LDR) to be used for loss calculation.

\section{Loss and Regularization Details}\label{append:loss}
This section presents the optimization objectives used in AEGIR. Inverse rendering is highly ill-posed, as different combinations of geometry, material, and illumination can produce the same image, so carefully designed losses are needed to guide the solution toward physically consistent results. This becomes even more challenging with explicit area emitters, which introduce additional degrees of freedom in shape, size, and emission, making optimization more difficult. The total optimization loss is defined in Equation \ref{eq:total_loss}.

\subsection{Contrast-Aware Photometric Loss}

Standard photometric losses often blur high-frequency details. We introduce a local contrast-aware focus map $W_f$ to prioritize the reconstruction of sharp lighting patterns. We define $\Delta I$ as the absolute difference between the grayscale ground-truth image $I_{gray}$ and its local average:
\begin{equation}
\Delta I = \left| I_{gray} - \text{AvgPool}(I_{gray}) \right|
\end{equation}
\begin{equation}
W_f = 1 + \lambda_{focus} \frac{\Delta I}{\max(\Delta I) + \epsilon}
\end{equation}
The final objective combines this weighted L1 loss with the Structural Similarity Index Measure (SSIM):
\begin{equation}
\mathcal{L}_{photo} = \| W_f \odot (\hat{I} - I) \|_1 + \lambda_{dssim} (1 - \text{SSIM}(\hat{I}, I))
\end{equation}
where $\hat{I}$ is the rendered image and $I$ is the ground-truth image.

\subsection{Material Supervision with Diffusion Priors}

To resolve ambiguities, such as distinguishing a white wall in shadow from a dark wall, we use pseudo-ground truth maps ($P_{prior}$) for base color $\mathbf{c}$, roughness $r$, and metallic $m$ extracted using DiffusionRenderer \cite{DiffusionRenderer}. We apply L1 regularization between our rendered G-buffer maps and these priors:
\begin{equation}
\mathcal{L}_{prior} = \sum_{P \in \{\mathbf{c}, r, m\}} \lambda_P \| P_{render} - P_{prior} \|_1
\end{equation}

\subsection{Edge-Aware Material Total Variation (TV)}

To mitigate multi-view noise introduced by the diffusion priors, we apply an edge-aware Total Variation (TV) loss. This loss encourages materials to remain constant across flat surfaces while preserving sharp transitions at structural boundaries. By smoothing the material textures, we prevent them from absorbing illumination gradients, which forces the light emitters to correctly account for these lighting effects:
\begin{equation}
\mathcal{L}_{TV} = \lambda_{TV} \sum_{P \in \{\mathbf{c}, r, m\}} \gamma_P \sum_{a,b} \|\nabla P_{a,b}\| \exp \left( - \beta \|\nabla I_{a,b}\| \right)
\end{equation}
where $\nabla P_{a,b}$ represents the spatial gradient of the material parameter $P$ at pixel $(a,b)$, $\gamma_P$ is the material-specific TV weight, and $\beta$ controls the sensitivity to the visual edges $\nabla I$ in the ground-truth image.

\subsection{Area Emitter Regularization}

To keep the lighting realistic, we apply three constraints to the explicit emitters. First, an area loss ($\mathcal{L}_{area}$) keeps the emitters from growing too large by penalizing the scale $s_i$ of each of the $N$ emitters:
\begin{equation}
\mathcal{L}_{area} = \frac{1}{N} \sum_{i=1}^{N} \|s_i\|^2
\end{equation}

Second, a soft white-light prior ($\mathcal{L}_{white}$) penalizes differences across the color channels $c$ of the emitter intensity $E_i$. This prevents the optimization from explaining material colors with unrealistic lighting tints, while still allowing the recovery of actual colored light sources:
\begin{equation}
\mathcal{L}_{white} = \frac{1}{N} \sum_{i=1}^{N} \left( \max_c (E_{i,c}) - \min_c (E_{i,c}) \right)
\end{equation}

Third, a bounding loss ($\mathcal{L}_{bounds}$) keeps the emitters inside the scene. It ensures the position $p_i$ of each emitter remains within the allowed boundaries $[b_{min}, b_{max}]$ across all spatial dimensions $d \in \{x,y,z\}$:
\begin{equation}
\mathcal{L}_{bounds} = \frac{1}{N} \sum_{i=1}^{N} \sum_{d \in \{x,y,z\}} \left( \max(0, b_{min,d} - p_{i,d}) + \max(0, p_{i,d} - b_{max,d}) \right)
\end{equation}

Finally, the total lighting regularization loss is a weighted sum of these three constraints:
\begin{equation}
\mathcal{L}_{light} = \lambda_{area} \mathcal{L}_{area} + \lambda_{white} \mathcal{L}_{white} + \lambda_{bounds} \mathcal{L}_{bounds}
\end{equation}

\section{Implementation and Optimization Details}\label{append:implement}
This section details the experimental setup used to evaluate the AEGIR pipeline, including the selected datasets, evaluation metrics, training schedules, dynamic emitter control strategies, and hyperparameter configurations, to support reproducibility.

\paragraph{Datasets and Splits.} We evaluate our method on a diverse collection of synthetic environments and real-world indoor scans. For synthetic data, we use the FIPT \cite{fipt2023} dataset, which contains 4 synthetic and 2 real-world scenes, and select 6 scenes from Hypersim \cite{roberts:2021} featuring complex indoor illumination patterns: \texttt{ai\_001\_004}, \texttt{ai\_004\_002}, \texttt{ai\_010\_006}, \texttt{ai\_015\_003}, \texttt{ai\_032\_001}, and \texttt{ai\_034\_002}. For real-world data, we evaluate on 8 indoor scans from the Replica \cite{replica19arxiv} dataset and 4 indoor scenes from ScanNet++ \cite{yeshwanth2023scannet++}: \texttt{0a7cc12c0e}, \texttt{8b5caf3398}, \texttt{a24858e51e}, and \texttt{be0ed6b33c}. Additionally, for our explicit lighting evaluation, we use the same synthetic scenes used in FIPT, which are originally from Benedikt Bitterli's rendering resources \cite{resources16}. Table \ref{tab:dataset_protocol} summarizes the shared evaluation protocol used across methods.

\begin{table}[h]
    \centering
    \caption{\textbf{Evaluation protocol.} All methods are evaluated on the same train/test camera splits for each dataset.}
    \label{tab:dataset_protocol}
    \vspace{1mm}
    \resizebox{\linewidth}{!}{
    \begin{tabular}{l c l l}
        \toprule
        Dataset & Scenes & Scene identifiers / source & Evaluation target \\
        \midrule
        Hypersim & 6 & \texttt{ai\_001\_004}, \texttt{ai\_004\_002}, \texttt{ai\_010\_006}, \texttt{ai\_015\_003}, \texttt{ai\_032\_001}, \texttt{ai\_034\_002} & NVS, albedo \\
        FIPT-Synthetic & 4 & FIPT synthetic split & NVS, albedo, roughness, ablations \\
        FIPT-Real & 2 & FIPT real split & NVS \\
        Replica & 8 & Replica indoor scans & NVS, ablations \\
        ScanNet++ & 4 & \texttt{0a7cc12c0e}, \texttt{8b5caf3398}, \texttt{a24858e51e}, \texttt{be0ed6b33c} & NVS \\
        Bitterli's Scenes & 4 & FIPT synthetic lighting scenes & Mitsuba lighting-only evaluation \\
        \bottomrule
    \end{tabular}
    }
\end{table}

\paragraph{Metrics.} To evaluate output quality, we use Peak Signal-to-Noise Ratio (PSNR) for pixel-wise accuracy, the Structural Similarity Index Measure (SSIM) for structural fidelity, and Learned Perceptual Image Patch Similarity (LPIPS) for deep perceptual similarity \cite{zhang2018perceptual}.

\paragraph{Initialization and Training Setup.} We initialize the scene geometry from a DN-Splatter \cite{turkulainen2024dnsplatter} checkpoint pretrained for 10,000 iterations. For physically based materials, all 2D Gaussians are initialized with a roughness of $0.6$ and a metallic value of $0.2$. Following this initialization, we run our deferred shading optimization for a total of 20,000 steps using the Adam optimizer. The initial learning rates are set to $1 \times 10^{-3}$ for material and lighting parameters, and $1 \times 10^{-5}$ for geometry. Between steps 5,000 and 15,000, the learning rates for geometry and materials decay to zero, effectively freezing them. During this same period, the emitter learning rate decays to $1 \times 10^{-4}$, allowing the lighting to continue refining exclusively for the final 5,000 steps.

For rendering, we employ the 2D Gaussian tracer from IRGS \cite{gu2024IRGS}, implemented in OptiX \cite{10.1145/1778765.1778803}. To support dynamic geometry updates, the bounding volume hierarchy (BVH) is rebuilt at every iteration, requiring about $3$ ms. Ray marching terminates when transmittance falls below $0.05$. During optimization, we sample $2^{16}$ rays per iteration. For the rendering equation evaluation, we use $N_{\text{brdf}}=128$ samples for diffuse transport and $N_{\text{light}}=128$ samples for explicit area emitters. The SH feature query is used only during training as a proxy for the single secondary bounce; final inference disables this proxy and evaluates the bounce by diffuse sampling and tracing to the recovered emitters.

In total, training takes about $55$ minutes on a single NVIDIA RTX 4090 GPU: $15$ minutes for the geometry initialization and $40$ minutes for the deferred shading optimization. For a fair comparison, AEGIR and our GS-ID reproduction use the same DN-Splatter geometry initialization. For baselines whose public implementations require a different scene representation or geometry pipeline, we follow the authors' recommended setup and evaluate on the same train/test views.

\paragraph{Adaptive Emitter Control.} To efficiently represent complex lighting without excessive memory consumption or over-exposure, we dynamically adapt the set of area emitters through splitting and pruning. These updates occur every 1,000 iterations between training steps 5,000 and 15,000. We evaluate each emitter $e$ based on its parameters, including scale $\mathbf{S}_e$, center $p_e$, and a perceptual energy measure $\Phi_e$. Given a linear RGB emission $\mathbf{E}_e=(E_{e,R}, E_{e,G}, E_{e,B})$, this energy is defined as:
\begin{equation}
\Phi_e = \left( \frac{E_{e,R} + E_{e,G} + E_{e,B}}{3} \right)^{1/2.2}.
\label{eq:app_emitter_energy}
\end{equation}

\begin{itemize}
    \item \textbf{Pruning:} We remove an emitter if its contribution becomes negligible ($\Phi_e < 0.1$) or physically implausible (drifting outside valid scene bounds).
    
    \item \textbf{Splitting:} To increase sampling resolution around dominant light sources, we split an emitter if it becomes excessively bright ($\Phi_e > 5.0$) or if its largest scale exceeds $20\%$ of the maximum scene scale. The emitter is divided along its major axis into two children, with each inheriting half the intensity and half the spatial scale of the parent.
\end{itemize}

\paragraph{Regularization Weights.} We balance accurate image reconstruction with realistic physics using the following tuned loss weights:

\begin{itemize}
    \item \textbf{Photometric Loss:} We apply a focus multiplier of $\lambda_{focus} = 5.0$ to the local contrast within the image reconstruction loss, and set the structural dissimilarity weight to $\lambda_{dssim} = 0.2$.
    \item \textbf{Material Regularization:} For the material guidance from the pretrained DiffusionRenderer, we apply equal weighting to the base color, roughness, and metallic components ($\lambda_{\mathbf{c}} = \lambda_r = \lambda_m = 1.0$). We regularize these materials using a total variation (TV) loss ($\lambda_{TV} = 1.0$), scaling the individual components by $\gamma_{\mathbf{c}} = 0.5$, $\gamma_r = 0.25$, and $\gamma_m = 0.25$, respectively. Finally, we set $\beta = 1.0$.
    \item \textbf{Emitter Regularization:} The light regularization terms $\lambda_{area}$ and $\lambda_{bounds}$ use a weight of $0.5$. However, $\lambda_{white}$ is set to a lower value of $0.1$ to allow the optimization enough flexibility to recover colored light sources.
\end{itemize}

\paragraph{Statistical Significance.} Due to the significant computational cost of optimization-based inverse rendering, we follow standard practices in neural rendering and report metrics from single optimization runs. However, to evaluate optimization stability, we conduct multiple runs with different seeds on the four synthetic scenes in the FIPT dataset and observe low standard deviations across seeds ($\pm0.14$ PSNR, $\pm0.014$ SSIM, $\pm0.022$ LPIPS).

\section{Additional Qualitative Results}\label{append:visual}
\begin{figure}[H]
    \centering
    \renewcommand{\arraystretch}{0.72}
    
    \def\envW{0.24\textwidth}  
    \newlength{\envH}
    \setlength{\envH}{0.12\textwidth} 
    \def\envGap{1.4mm} 
    
    \newcommand{\EnvImg}[1]{%
      \setlength{\fboxsep}{0pt}%
      \setlength{\fboxrule}{0.5pt}%
      \fbox{%
        \parbox[c][\dimexpr\envH-2\fboxrule\relax][c]{\dimexpr\envW-2\fboxrule\relax}{%
          \centering
          \IfFileExists{#1}{%
            \includegraphics[width=\dimexpr\envW-2\fboxrule\relax, height=\dimexpr\envH-2\fboxrule\relax, keepaspectratio]{#1}%
          }{%
            \textbf{\small N/A}%
          }%
        }%
      }%
    }

    \newcommand{\EnvImgDouble}[1]{%
      \setlength{\fboxsep}{0pt}%
      \setlength{\fboxrule}{0.5pt}%
      \fbox{%
        \parbox[c][\dimexpr 2\envH + \envGap - 2\fboxrule\relax][c]{\dimexpr\envW-2\fboxrule\relax}{%
          \centering
          \IfFileExists{#1}{%
            \includegraphics[width=\dimexpr\envW-2\fboxrule\relax, height=\dimexpr 2\envH + \envGap - 2\fboxrule\relax, keepaspectratio]{#1}%
          }{%
            \textbf{\small N/A}%
          }%
        }%
      }%
    }
    
    \newcommand{\EnvStacked}[2]{%
      \parbox[c][\dimexpr 2\envH + \envGap \relax][c]{\envW}{%
        \offinterlineskip 
        \EnvImg{#1}\par
        \vspace{\envGap}%
        \EnvImg{#2}%
      }%
    }
    
    \setlength{\tabcolsep}{1.4pt} 
    \begin{tabular}{@{} *{4}{c} @{}}
        \small Reference View & 
        \small GT Environment Map & 
        \small \textbf{AEGIR (ours)} & 
        \small LuxDiT \cite{liang2025luxdit}\\[1mm]

        \EnvImgDouble{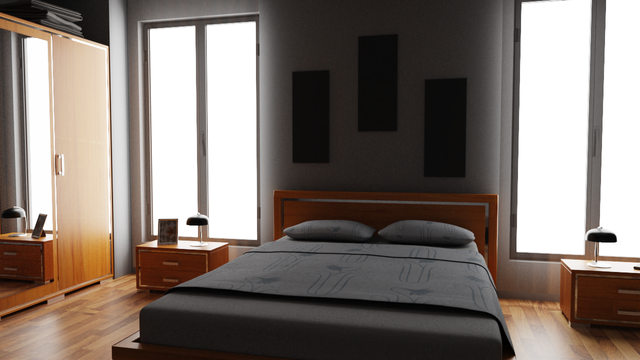} &
        \EnvStacked{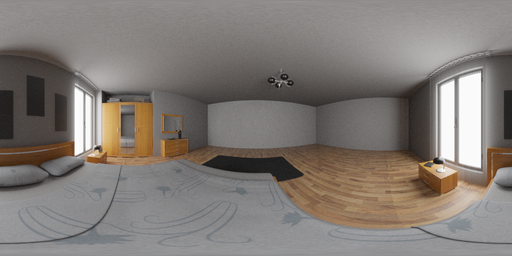}{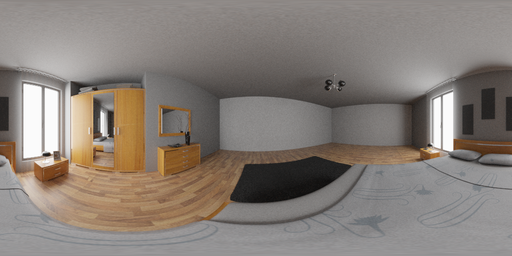} &
        \EnvStacked{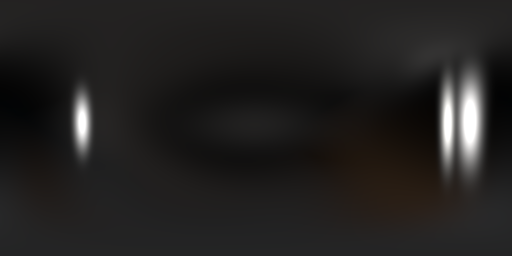}{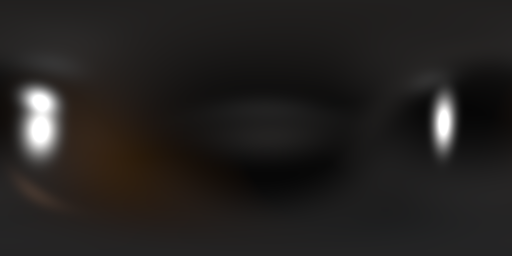} &
        \EnvStacked{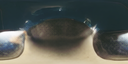}{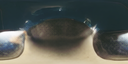} \\[\envGap]

        \EnvImgDouble{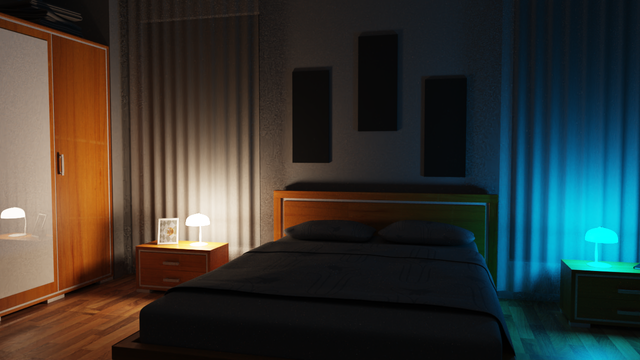} &
        \EnvStacked{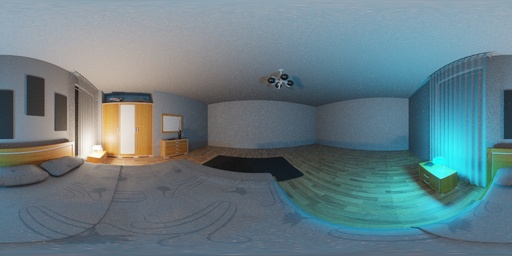}{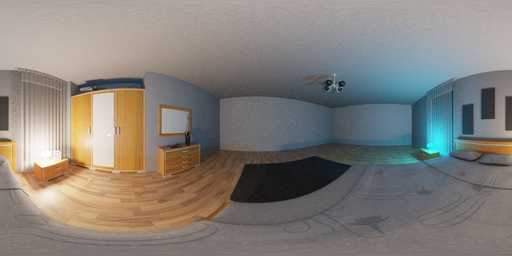} &
        \EnvStacked{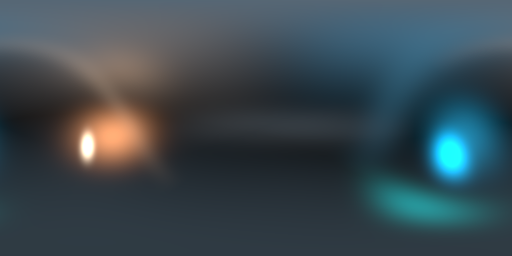}{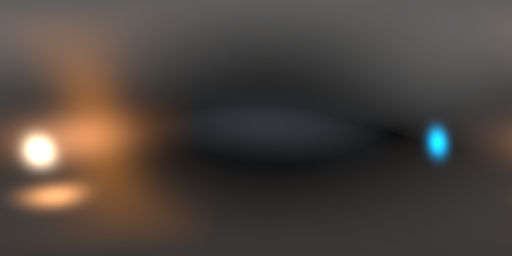} &
        \EnvStacked{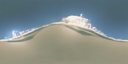}{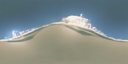} \\
    \end{tabular}

    \vspace{1mm}
    \caption{\textbf{Qualitative environment map comparison.} Environment maps are extracted at two spatial locations per lighting condition. LuxDiT \cite{liang2025luxdit} does not produce spatially varying maps, failing to capture localized lighting. AEGIR successfully models this spatial variation and anisotropic light sources, closely matching the ground truth.}
    \label{fig:envmap_spatial_points}
\end{figure}

\begin{figure}[t]
    \centering
    \renewcommand{\arraystretch}{0.72}
    \def\illumW{0.285\textwidth}  
    \newlength{\illumH}
    \setlength{\illumH}{0.18\textwidth} 
    
    \newcommand{\IllumImg}[1]{%
      \setlength{\fboxsep}{0pt}%
      \setlength{\fboxrule}{0.5pt}%
      \fbox{%
        \parbox[c][\dimexpr\illumH-2\fboxrule\relax][c]{\dimexpr\illumW-2\fboxrule\relax}{%
          \centering
          \IfFileExists{#1}{%
            \includegraphics[width=\dimexpr\illumW-2\fboxrule\relax, height=\dimexpr\illumH-2\fboxrule\relax, keepaspectratio]{#1}%
          }{%
            \textbf{\small N/A}%
          }%
        }%
      }%
    }
    
    \setlength{\tabcolsep}{1.4pt}
    \begin{tabular}{@{} *{3}{c} @{}}
        \small Reference & 
        \small \textbf{AEGIR} & 
        \small GS-ID \cite{du2025gsidilluminationdecompositiongaussian} \\[1mm] 

        \IllumImg{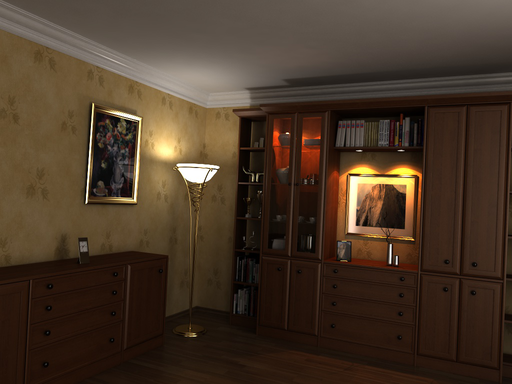} &
        \IllumImg{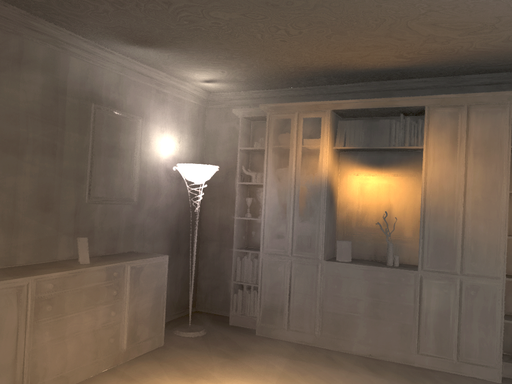} &
        \IllumImg{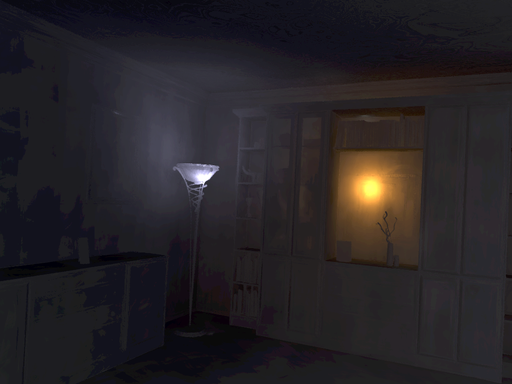} \\[1.2mm]

        \IllumImg{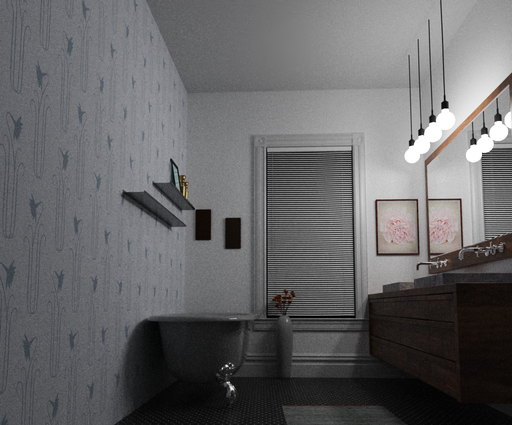} &
        \IllumImg{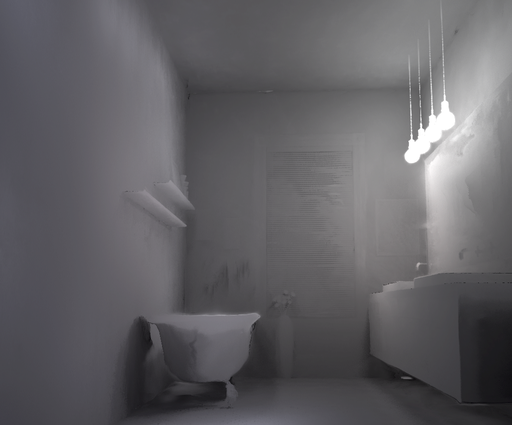} &
        \IllumImg{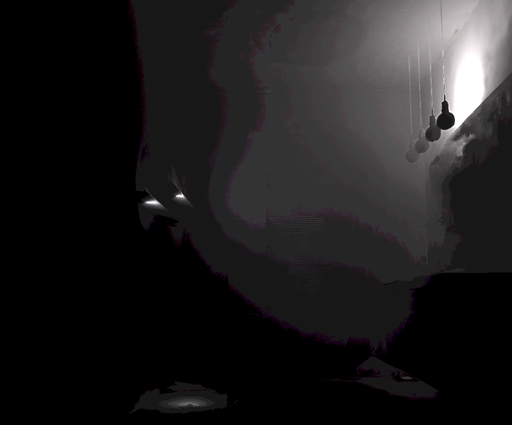} \\[1.2mm]

        \IllumImg{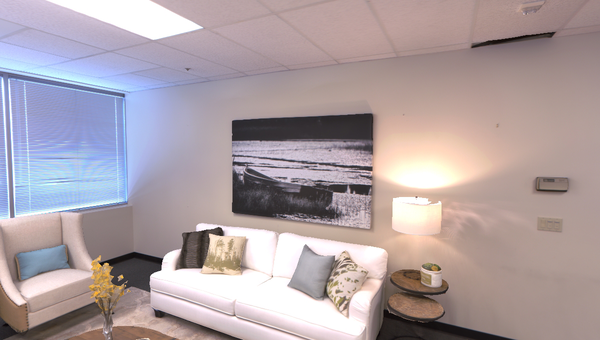} &
        \IllumImg{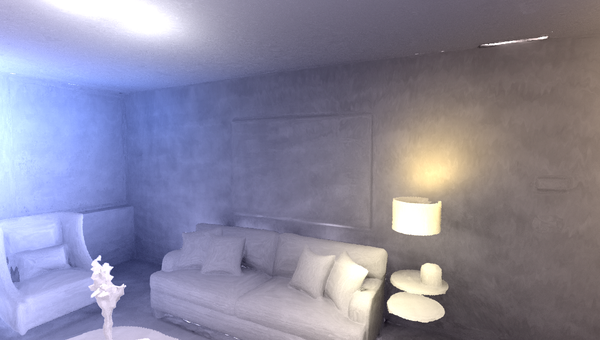} &
        \IllumImg{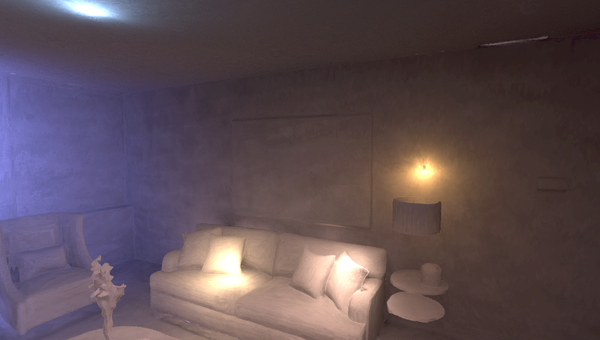} \\[1.2mm]

        \IllumImg{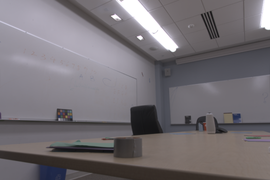} &
        \IllumImg{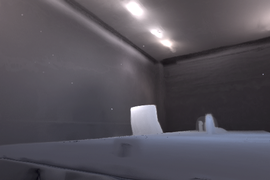} &
        \IllumImg{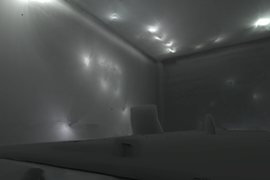} \\
    \end{tabular}

    \vspace{1.5mm}
    \caption{\textbf{Qualitative comparison of illumination representations.} Unlike GS-ID \cite{du2025gsidilluminationdecompositiongaussian}, which relies on discrete point lights, AEGIR's explicit area emitters accurately recover the spatial extent and physical structure of light sources in direct illumination (shading) maps.}
    \label{fig:illumination_emitters}

\end{figure}

\begin{figure*}[p]
    \centering
    
    \newcommand{\SideLabel}[1]{$\vcenter{\hbox{\rotatebox{90}{\scriptsize \textbf{#1}}}}$}
    
    \newcommand{\ResImg}[1]{%
        $\vcenter{\hbox{%
        \parbox[c]{0.130\textwidth}{%
            \centering
            \IfFileExists{#1}{%
                \includegraphics[width=0.130\textwidth, keepaspectratio]{#1}%
            }{%
                \vspace{2mm}\small\textbf{N/A}\vspace{2mm}%
            }%
        }%
        }}$%
    }
    
    \newcommand{\SceneBlock}[2]{%
        $\vcenter{\hbox{%
        \renewcommand{\arraystretch}{0.68}%
        \begin{tabular}{@{} c @{\hspace{1.35mm}} c @{\hspace{2.7mm}} c @{\hspace{1.35mm}} c @{\hspace{1.35mm}} c @{\hspace{1.35mm}} c @{\hspace{1.35mm}} c @{}}
            \SideLabel{Render} &
            \ResImg{#2} &
            \ResImg{figures/nvs/#1/aegir/rgb.png} &
            \ResImg{figures/nvs/#1/gsid/rgb.png} &
            \ResImg{figures/nvs/#1/irgs/rgb.png} &
            \ResImg{figures/nvs/#1/iris/rgb.png} &
            \ResImg{figures/nvs/#1/neilfpp/rgb.png} \\[1.7mm]
            
            \SideLabel{Albedo} &
            \ResImg{figures/nvs/#1/albedo.png} &
            \ResImg{figures/nvs/#1/aegir/albedo.png} &
            \ResImg{figures/nvs/#1/gsid/albedo.png} &
            \ResImg{figures/nvs/#1/irgs/albedo.png} &
            \ResImg{figures/nvs/#1/iris/albedo.png} &
            \ResImg{figures/nvs/#1/neilfpp/albedo.png} \\[1.7mm]
            
            \SideLabel{Roughness} &
            \ResImg{figures/nvs/#1/rough.png} &
            \ResImg{figures/nvs/#1/aegir/rough.png} &
            \ResImg{figures/nvs/#1/gsid/rough.png} &
            \ResImg{figures/nvs/#1/irgs/rough.png} &
            \ResImg{figures/nvs/#1/iris/rough.png} &
            \ResImg{figures/nvs/#1/neilfpp/rough.png} \\[1.7mm]
            
            \SideLabel{Metallic} &
            \ResImg{figures/nvs/#1/metal.png} &
            \ResImg{figures/nvs/#1/aegir/metal.png} &
            \ResImg{figures/nvs/#1/gsid/metal.png} &
            \ResImg{figures/nvs/#1/irgs/metal.png} &
            \ResImg{figures/nvs/#1/iris/metal.png} &
            \ResImg{figures/nvs/#1/neilfpp/metal.png} \\
        \end{tabular}%
        }}$ \\
    }
    
    \begin{tabular}{@{} c @{}}
        
        \begin{tabular}{@{} c @{\hspace{1.35mm}} c @{\hspace{2.7mm}} c @{\hspace{1.35mm}} c @{\hspace{1.35mm}} c @{\hspace{1.35mm}} c @{\hspace{1.35mm}} c @{}}
            \makebox[8pt]{} &
            \parbox[c]{0.130\textwidth}{\centering \small Reference} &
            \parbox[c]{0.130\textwidth}{\centering \small \textbf{AEGIR (Ours)}} &
            \parbox[c]{0.130\textwidth}{\centering \small GS-ID \cite{du2025gsidilluminationdecompositiongaussian}} &
            \parbox[c]{0.130\textwidth}{\centering \small IRGS \cite{gu2024IRGS}} &
            \parbox[c]{0.130\textwidth}{\centering \small IRIS \cite{lin2025iris}} &
            \parbox[c]{0.130\textwidth}{\centering \small NeILF++ \cite{zhang2023neilf++}} \\
        \end{tabular} \\
        
        \toprule[0.8pt]
        
        \SceneBlock{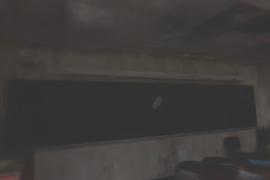}{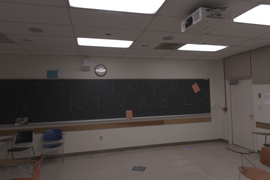}
        \midrule
        
        \SceneBlock{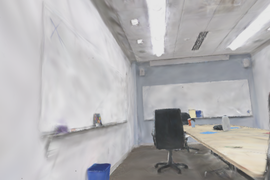}{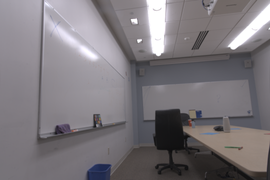}
        \midrule
        
        \SceneBlock{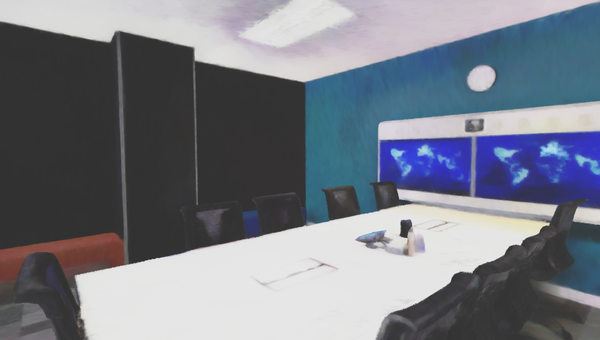}{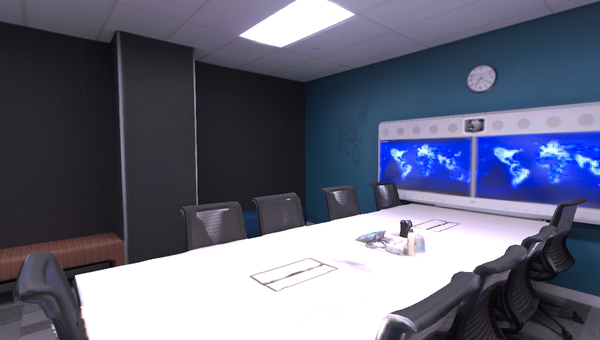}
        \bottomrule[0.8pt]
        
    \end{tabular}
    
    \vspace{2mm}
    \caption{\textbf{Additional qualitative comparison of novel view synthesis and intrinsic decomposition on indoor scenes.} The reference column follows the same row layout as the method outputs, showing the render/RGB image, albedo, roughness, and metallic maps when available. The remaining columns compare the estimated properties and renderings across methods. ``N/A'' indicates properties not produced by a given method.}
    \label{fig:intrinsic_layout_appendix}
\end{figure*}

\begin{figure*}[p]
    \centering
    
    \newcommand{\NewSceneSideLabel}[1]{$\vcenter{\hbox{\rotatebox{90}{\scriptsize \textbf{#1}}}}$}
    
    \newcommand{\NewSceneResImg}[1]{%
        $\vcenter{\hbox{%
        \parbox[c]{0.130\textwidth}{%
            \centering
            \IfFileExists{#1}{%
                \includegraphics[width=0.130\textwidth, keepaspectratio]{#1}%
            }{%
                \vspace{2mm}\small\textbf{N/A}\vspace{2mm}%
            }%
        }%
        }}$%
    }
    
    \newcommand{\NewSceneBlock}[2]{%
        $\vcenter{\hbox{%
        \renewcommand{\arraystretch}{0.68}%
        \begin{tabular}{@{} c @{\hspace{1.35mm}} c @{\hspace{2.7mm}} c @{\hspace{1.35mm}} c @{\hspace{1.35mm}} c @{\hspace{1.35mm}} c @{\hspace{1.35mm}} c @{}}
            \NewSceneSideLabel{Render} &
            \NewSceneResImg{#2} &
            \NewSceneResImg{figures/nvs/#1/aegir/rgb.png} &
            \NewSceneResImg{figures/nvs/#1/gsid/rgb.png} &
            \NewSceneResImg{figures/nvs/#1/irgs/rgb.png} &
            \NewSceneResImg{figures/nvs/#1/iris/rgb.png} &
            \NewSceneResImg{figures/nvs/#1/neilfpp/rgb.png} \\[1.7mm]
            
            \NewSceneSideLabel{Albedo} &
            \NewSceneResImg{figures/nvs/#1/albedo.png} &
            \NewSceneResImg{figures/nvs/#1/aegir/albedo.png} &
            \NewSceneResImg{figures/nvs/#1/gsid/albedo.png} &
            \NewSceneResImg{figures/nvs/#1/irgs/albedo.png} &
            \NewSceneResImg{figures/nvs/#1/iris/albedo.png} &
            \NewSceneResImg{figures/nvs/#1/neilfpp/albedo.png} \\[1.7mm]
            
            \NewSceneSideLabel{Roughness} &
            \NewSceneResImg{figures/nvs/#1/rough.png} &
            \NewSceneResImg{figures/nvs/#1/aegir/rough.png} &
            \NewSceneResImg{figures/nvs/#1/gsid/rough.png} &
            \NewSceneResImg{figures/nvs/#1/irgs/rough.png} &
            \NewSceneResImg{figures/nvs/#1/iris/rough.png} &
            \NewSceneResImg{figures/nvs/#1/neilfpp/rough.png} \\[1.7mm]
            
            \NewSceneSideLabel{Metallic} &
            \NewSceneResImg{figures/nvs/#1/metal.png} &
            \NewSceneResImg{figures/nvs/#1/aegir/metal.png} &
            \NewSceneResImg{figures/nvs/#1/gsid/metal.png} &
            \NewSceneResImg{figures/nvs/#1/irgs/metal.png} &
            \NewSceneResImg{figures/nvs/#1/iris/metal.png} &
            \NewSceneResImg{figures/nvs/#1/neilfpp/metal.png} \\
        \end{tabular}%
        }}$ \\
    }
    
    \begin{tabular}{@{} c @{}}
        \begin{tabular}{@{} c @{\hspace{1.35mm}} c @{\hspace{2.7mm}} c @{\hspace{1.35mm}} c @{\hspace{1.35mm}} c @{\hspace{1.35mm}} c @{\hspace{1.35mm}} c @{}}
            \makebox[8pt]{} &
            \parbox[c]{0.130\textwidth}{\centering \small Reference} &
            \parbox[c]{0.130\textwidth}{\centering \small \textbf{AEGIR (Ours)}} &
            \parbox[c]{0.130\textwidth}{\centering \small GS-ID \cite{du2025gsidilluminationdecompositiongaussian}} &
            \parbox[c]{0.130\textwidth}{\centering \small IRGS \cite{gu2024IRGS}} &
            \parbox[c]{0.130\textwidth}{\centering \small IRIS \cite{lin2025iris}} &
            \parbox[c]{0.130\textwidth}{\centering \small NeILF++ \cite{zhang2023neilf++}} \\
        \end{tabular} \\
        
        \toprule[0.8pt]
        \NewSceneBlock{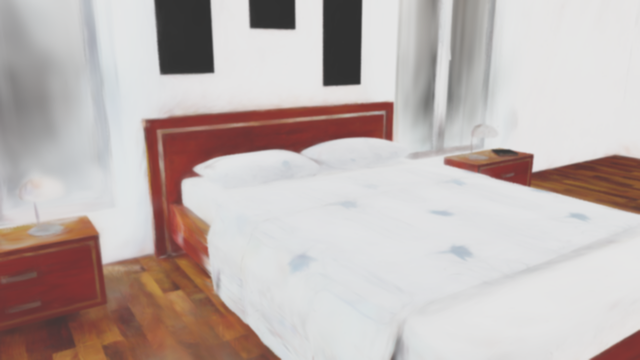}{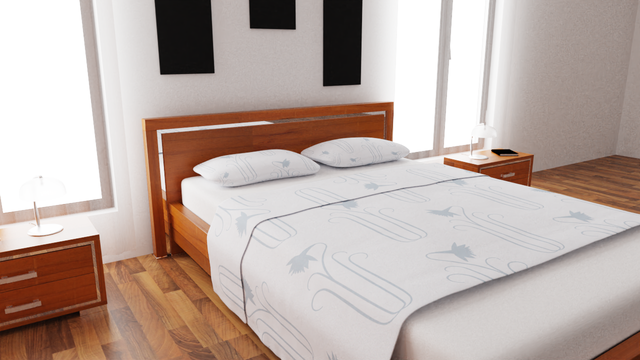}
        \midrule
        \NewSceneBlock{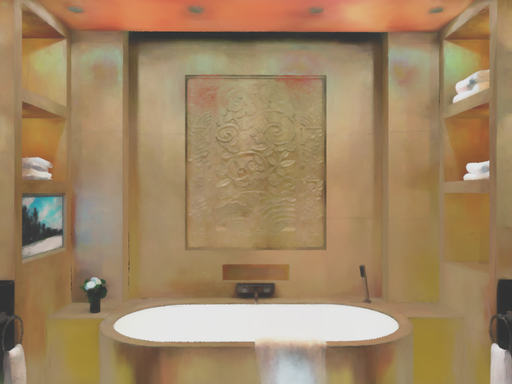}{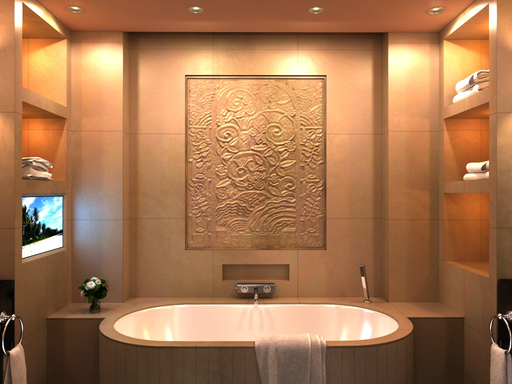}
        \midrule
        \NewSceneBlock{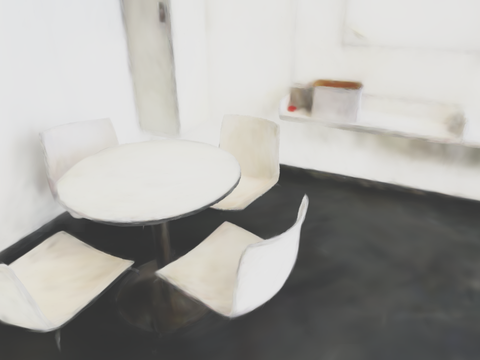}{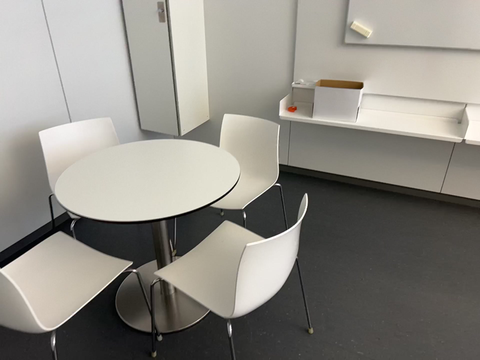}
        \bottomrule[0.8pt]
    \end{tabular}
    
    \vspace{2mm}
    \caption{\textbf{Additional qualitative comparison of novel view synthesis and intrinsic decomposition on indoor scenes.} This figure extends the appendix comparison with two newly added scenes and one additional real-world scan. The reference column follows the same row layout as the method outputs, showing the render/RGB image, albedo, roughness, and metallic maps when available. ``N/A'' indicates properties not produced by a given method.}
    \label{fig:intrinsic_layout_appendix_new_scenes}
\end{figure*}

\begin{figure*}[t]
    \centering
    
    \newcommand{\MVTitle}[1]{$\vcenter{\hbox{\rotatebox{90}{\small \textbf{#1}}}}$}
    
    \newcommand{\MVImg}[3]{%
      $\vcenter{\hbox{%
      \IfFileExists{#2}{%
        \includegraphics[width=#1]{#2}%
      }{%
        \setlength{\fboxsep}{0pt}%
        \setlength{\fboxrule}{0.4pt}%
        \fbox{\parbox[c][#3][c]{\dimexpr#1-2\fboxsep-2\fboxrule\relax}{\centering \textbf{\small N/A}}}%
      }%
      }}$%
    }
    
    \def\mvWidth{0.225\textwidth}
    \def\mvHeight{1.8cm} 

    \renewcommand{\arraystretch}{0.5} 
    \setlength{\tabcolsep}{1pt}        

    \begin{tabular}{@{} c @{\hspace{2mm}} *{4}{c} @{}}
        \MVTitle{\tiny GT RGB} &
        \MVImg{\mvWidth}{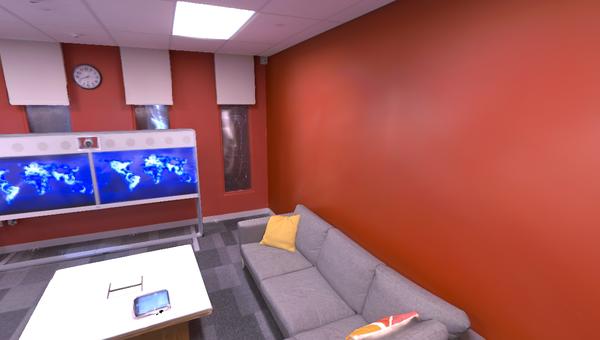}{\mvHeight} &
        \MVImg{\mvWidth}{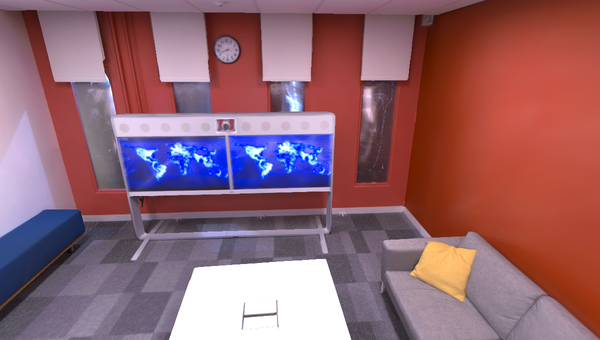}{\mvHeight} &
        \MVImg{\mvWidth}{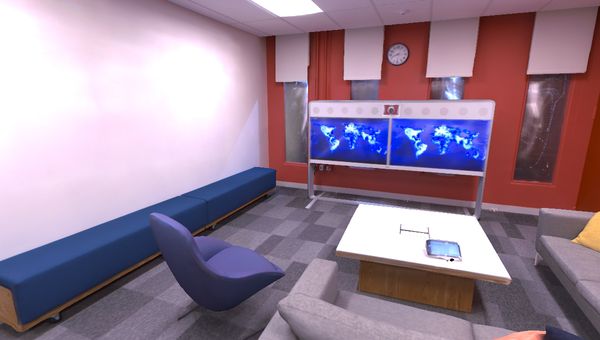}{\mvHeight} &
        \MVImg{\mvWidth}{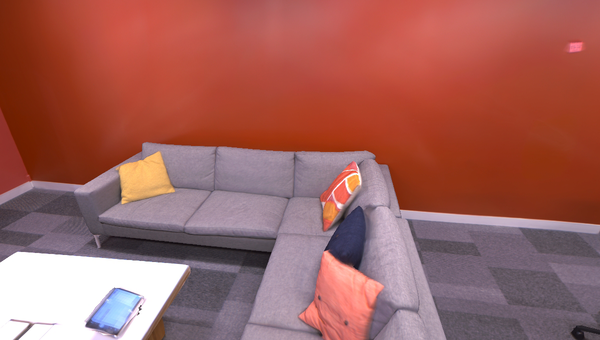}{\mvHeight} \\[1mm]
        
        \MVTitle{\tiny Diff. Rend. \cite{DiffusionRenderer}} &
        \MVImg{\mvWidth}{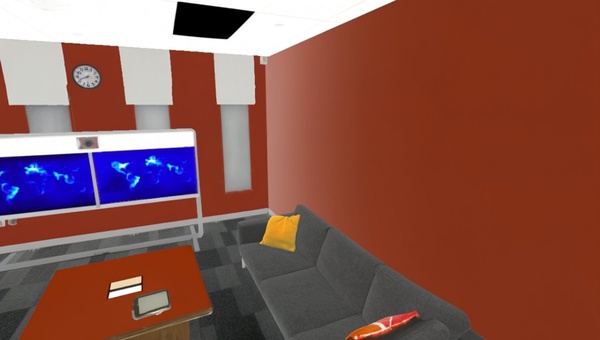}{\mvHeight} &
        \MVImg{\mvWidth}{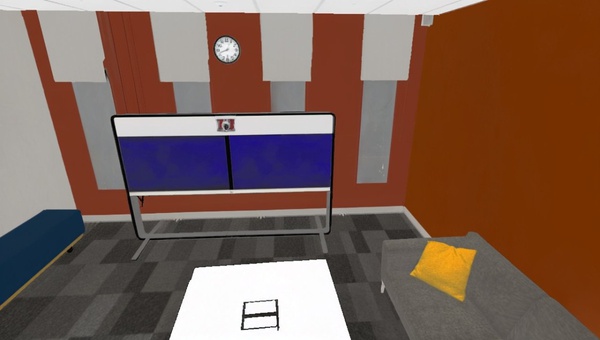}{\mvHeight} &
        \MVImg{\mvWidth}{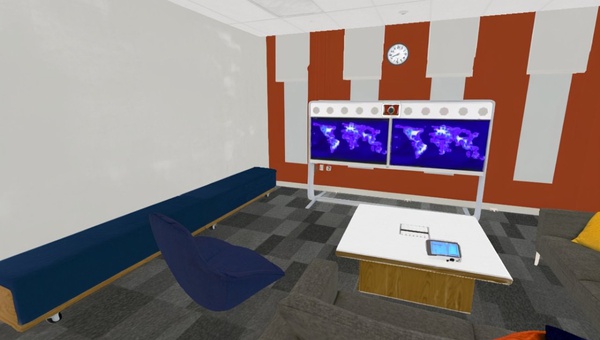}{\mvHeight} &
        \MVImg{\mvWidth}{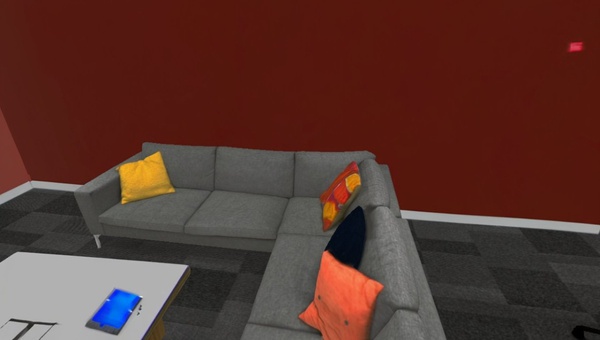}{\mvHeight} \\[1mm]
        
        \MVTitle{\tiny AEGIR (ours)} &
        \MVImg{\mvWidth}{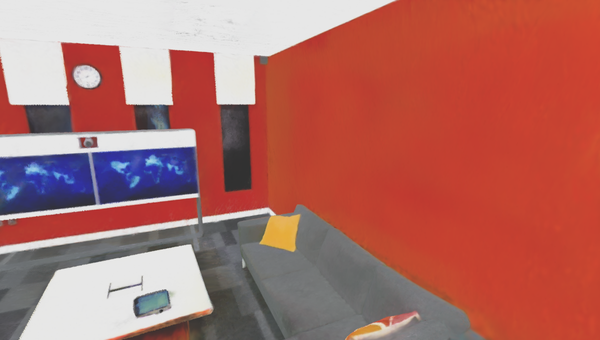}{\mvHeight} &
        \MVImg{\mvWidth}{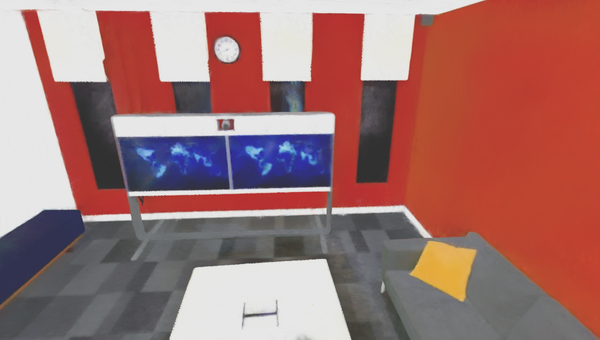}{\mvHeight} &
        \MVImg{\mvWidth}{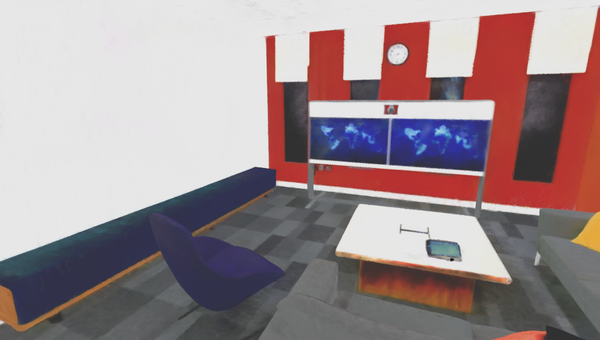}{\mvHeight} &
        \MVImg{\mvWidth}{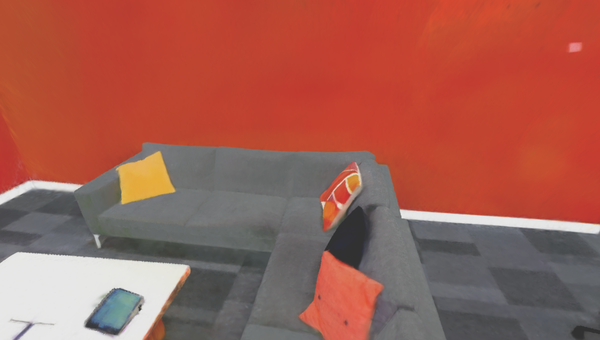}{\mvHeight} \\[10mm]
        
        \MVTitle{\tiny GT RGB} &
        \MVImg{\mvWidth}{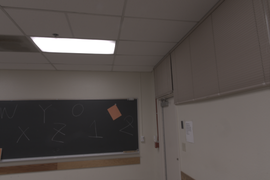}{\mvHeight} &
        \MVImg{\mvWidth}{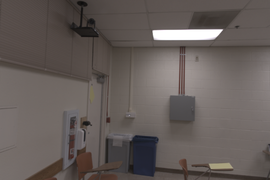}{\mvHeight} &
        \MVImg{\mvWidth}{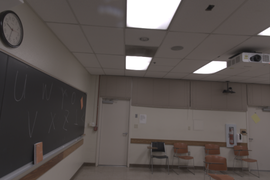}{\mvHeight} &
        \MVImg{\mvWidth}{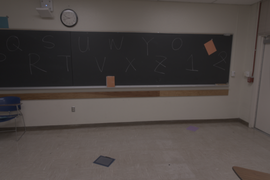}{\mvHeight} \\[1mm]
        
        \MVTitle{\tiny Diff. Rend. \cite{DiffusionRenderer}} &
        \MVImg{\mvWidth}{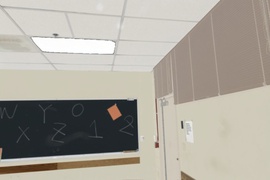}{\mvHeight} &
        \MVImg{\mvWidth}{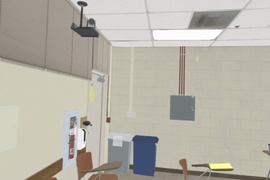}{\mvHeight} &
        \MVImg{\mvWidth}{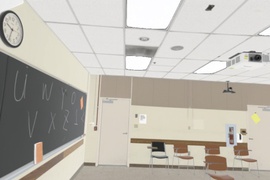}{\mvHeight} &
        \MVImg{\mvWidth}{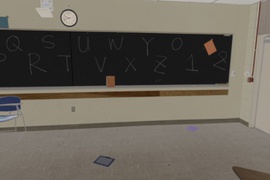}{\mvHeight} \\[1mm]
        
        \MVTitle{\tiny AEGIR (ours)} &
        \MVImg{\mvWidth}{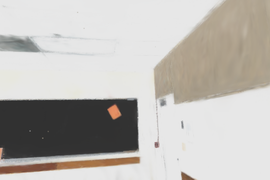}{\mvHeight} &
        \MVImg{\mvWidth}{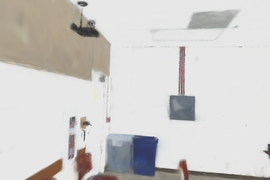}{\mvHeight} &
        \MVImg{\mvWidth}{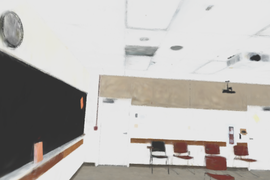}{\mvHeight} &
        \MVImg{\mvWidth}{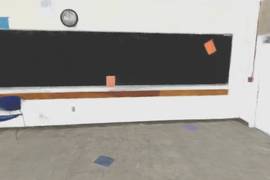}{\mvHeight} \\
    \end{tabular}

    \caption{\textbf{Material multi-view consistency.} Top to bottom: Ground-truth RGB, DiffusionRenderer \cite{DiffusionRenderer} albedo, and our AEGIR albedo. Unlike 2D generative priors that suffer from occasional illumination leakage and multi-view inconsistencies, AEGIR ensures robust and consistent material estimation across all viewpoints.}
    \label{fig:diffusion_renderer_multiview}
\end{figure*}

This section presents additional qualitative results that highlight the performance of AEGIR from multiple perspectives. The first two figures provide additional qualitative evaluations of lighting estimation. In Figure \ref{fig:envmap_spatial_points}, we render environment maps from the area emitters estimated by AEGIR at multiple spatial locations and compare them to ground-truth environment maps rendered in Mitsuba \cite{Mitsuba3}. This directly evaluates whether the recovered emitters reproduce the spatial variation of scene illumination. We also compare against LuxDiT \cite{liang2025luxdit}, a state-of-the-art diffusion-based method that estimates environment maps from single or multiple inputs. Despite using image sequences, LuxDiT predicts a single global environment map and therefore cannot capture spatial variation or localized lighting. In contrast, AEGIR more faithfully reproduces how the environment map changes across different spatial locations. Figure \ref{fig:illumination_emitters} provides a second qualitative lighting evaluation through direct illumination (shading) maps, showing that explicit area emitters recover the spatial extent of illumination more accurately than the discrete point light representation used in GS-ID \cite{du2025gsidilluminationdecompositiongaussian}. For material recovery, Figures \ref{fig:intrinsic_layout_appendix} and \ref{fig:intrinsic_layout_appendix_new_scenes} provide extended intrinsic evaluations, showing novel-view renderings and albedo, roughness, and metallic maps across real and synthetic indoor scans against baseline methods. Finally, Figure \ref{fig:diffusion_renderer_multiview} highlights the multi-view inconsistencies of 2D generative priors such as DiffusionRenderer \cite{DiffusionRenderer}, and shows how AEGIR produces more consistent material estimates.

\section{Limitations and Future Work}\label{append:limit}
Despite its strong performance in recovering physically grounded scene illumination, AEGIR still has several limitations that suggest directions for future work.

\textbf{Geometric limitations.} Our deferred shading pipeline relies on 2D Gaussian geometry. In textureless or heavily occluded regions, this can produce noisy or ambiguous surface normals, leading to visible artifacts during illumination. Future work could address this by incorporating stronger monocular geometric priors together with spatial smoothness constraints to improve normal stability in these regions.

\textbf{Scalability of explicit area light sampling.} The computational cost increases with the number of explicit area light sources. Since ray tracing and Monte Carlo integration scale with the emitter count, modeling complex illumination in large scenes can become a performance bottleneck. This could be addressed by integrating scalable light sampling methods such as ReSTIR \cite{10.1145/3386569.3392481}, which reuse samples over space and time to maintain low variance at lower cost.

\textbf{Limitations in transparent and specular materials.} Our opaque microfacet BRDF limits the accurate modeling of mirrors and transparent materials such as glass. Extending the representation to a full BSDF with transmission components, combined with differentiable refraction tracing, would enable physically accurate simulation of light transport through transmissive materials and expand the range of reconstructible scenes.


\end{document}